\documentclass[journal]{IEEEtran}
\IEEEpubid{\makebox[\columnwidth]{~
		\copyright2018
		IEEE \hfill} \hspace{\columnsep}\makebox[\columnwidth]{ }}
\usepackage{url}
\usepackage{graphicx}
\usepackage{multirow}
\usepackage[caption=false,font=footnotesize]{subfig}
\usepackage{amsmath}
\usepackage[group-separator={,}]{siunitx}
\usepackage{cite}
\usepackage{multirow}
\usepackage{epstopdf}
\usepackage{color}

\begin{document}
\title{Quality Diversity Through Surprise}	
\author{Daniele~Gravina, Antonios~Liapis~\IEEEmembership{Member,~IEEE}, and~Georgios~N. Yannakakis~\IEEEmembership{Senior Member,~IEEE}
\thanks{All authors are with the Institute of Digital Games, University of Malta, Msida 2080, Malta (e-mail: daniele.gravina@um.edu.mt; antonios.liapis@um.edu.mt; georgios.yannakakis@um.edu.mt)
}
}
\maketitle

\begin{abstract} 
Quality diversity is a recent family of evolutionary search algorithms which focus on finding several well-performing (\emph{quality}) yet different (\emph{diversity}) solutions with the aim to maintain an appropriate balance between divergence and convergence during search. While quality diversity has already delivered promising results in complex problems, the capacity of divergent search variants for quality diversity remains largely unexplored. Inspired by the notion of \emph{surprise} as an effective driver of divergent search and its orthogonal nature to \emph{novelty} this paper investigates the impact of the former to quality diversity performance. For that purpose we introduce three new quality diversity algorithms which employ surprise as a diversity measure, either on its own or combined with novelty, and compare their performance against novelty search with local competition, the state of the art quality diversity algorithm. The algorithms are tested in a robot navigation task across 60 highly deceptive mazes. Our findings suggest that allowing surprise and novelty to operate synergistically for divergence and in combination with local competition leads to quality diversity algorithms of significantly higher efficiency, speed and robustness.
\end{abstract}


\IEEEpeerreviewmaketitle

\section{Introduction}\label{sec:introduction} 

\vspace{1em}

\IEEEPARstart{Q}{uality} diversity algorithms have been recently introduced to the evolutionary computation (EC) literature as a way of handling deceptive search spaces. The goal of these algorithms is ``to find a maximally diverse collection of individuals (with respect to a space of possible behaviors) in which each member is as high performing as possible'' \cite{pugh2016quality}. As highlighted in \cite{pugh2016quality}, the inspiration for such approaches is natural evolution which is primarily open-ended---unlike the objective-based optimization tasks to which EC is often applied. While the rationale of open-ended evolution has been previously used as an argument for genetic search for pure behavioral novelty \cite{lehman2011abandoning}, quality diversity algorithms re-introduce a notion of (localized) quality among individuals with the same behavioral characteristics. In the natural evolution analogy, quality diversity would be similar to a competition among creatures with the same \emph{behavioral} traits\footnote{Since diversity refers to behavioral niches, notions such as species may refer to genotypic similarities and are not used in this paper to avoid confusion. To make matters more complex, the term speciation \cite{stanley2002neat} in EC similarly operates on the genotype rather than the phenotype of individuals.}. For instance, small flying animals (bats, birds) compete for the same food supply, but do not compete with large terrestrial herbivores.
	
Quality diversity (QD) algorithms attempt to balance between their individuals' quality and their population's diversity. Quality can be assessed via an objective function, assuming a problem space where this is possible to compute. Diversity, on the other hand, can been assessed in different ways: for instance, MAP-elites \cite{mouret2015illuminating} compartmentalizes the search space beforehand based on two or more behavioral characterizations, while novelty search with local competition \cite{lehman2011creatures,pugh2016quality} pushes for \emph{novelty} as a second objective. Novelty in this algorithm is assessed as the distance from the behaviorally closest neighbors in the current population and in an archive of past novel individuals. In that sense, novelty is the deviation from current and past solutions. In natural evolution, the novelty of a behavioral trait such as flying, coupled with a competitive advantage (such as the improved flying ability of Archaeopteryx against flying reptiles of the same period) can lead to large shifts as these behaviors become dominant.
	
Novelty, however, faces a number of limitations as a measure of diversity; in particular, it lacks a \emph{temporal} dimension in terms of the trends that evolution is following from one generation to the next. Towards that end, \emph{surprise} has been introduced as an alternative to novelty for divergent search \cite{yannakakis2016searching,gravina2016surprisebeyond}. Unlike novelty, surprise accounts for behavioral trends in recent generations and uses them to predict future behaviors: if a new individual breaks those expectations, then its behavior is surprising and the individual is favored for evolution. While surprise is tied to human emotions \cite{ekman1992argument}, in the context of EC it is more broadly defined as \emph{deviation from expected behaviors}. Based on that definition, surprise can be considered both orthogonal and complementary to novelty: the latter deviates from past behaviors, while the former deviates from predicted future behaviors. Both novelty and surprise have been applied as measures of divergence in EC, through novelty search \cite{lehman2011abandoning,mouret2011novelty} and surprise search \cite{gravina2016surprisebeyond} respectively. Both of these algorithms perform pure divergent search which has been shown to outperform objective search in deceptive environments such as maze navigation \cite{lehman2008exploiting,gravina2016surprisebeyond}, creature locomotion \cite{lehman2011abandoning,gravina2017soft}, or game content generation \cite{liapis2016arcade,gravina2016constrained}. Moreover, experiments have shown that surprise search discovers different niches of the search space than novelty search \cite{gravina2016constrained,gravina2017soft} and often results in more efficient and more robust evolutionary runs in more deceptive tasks \cite{gravina2016surprisebeyond}. 
	
Drawing an evolutionary analogy for novelty search and surprise search, one may view those processes through the lens of \emph{phylogenetics} \cite{wiley2011phylogenetics}; the study of the evolutionary history and relationships among organisms. Both algorithms may represent evolutionary lineages that operate synergistically on the behavioral (rather than on the genetic) space of a phylogenetic tree. While both processes can be seen as \emph{behavioral lineages} of evolution, on the one hand novelty search rewards diversity by aggregating the entire evolutionary history into a novelty archive, on the other hand surprise search considers the recent historical trends to make predictions about the future and deviate from them. Drawing again from natural evolution in paleontology, the shift towards smaller, flying dinosaurs in the Cretaceous period \cite{lee2016dinos} pointed to a trend towards ever-smaller fliers, as evident in most birds in following epochs: such birds would not be deemed surprising. On the other hand, some birds evolved into man-sized flightless bipedal predators \cite{bertelli2007kelenken} (``terror birds''): this breaks expectations and phenetic tendencies, therefore terror birds are deemed surprising. However, terror birds are not behaviorally novel since similar-sized bipedal predators abounded in earlier periods (i.e., most carnivorous dinosaurs) \cite{paul1988predatory}.

With the theoretical argument for the orthogonal nature of novelty and surprise \cite{yannakakis2016searching} and the promising results of the latter for divergent search \cite{gravina2016surprisebeyond,gravina2018fusing}, this paper introduces three new quality diversity algorithms relying on surprise. Inspired by novelty search with local competition (NS-LC), the new algorithms replace novelty search with surprise search (SS-LC), or combine measures of novelty and surprise linearly (NSS-LC) or as separate objectives (NS-SS-LC). We compare the performance of these algorithms in solving robot navigation tasks across 60 different generated mazes in which controllers of wheeled robots are evolved via neuroevolution of augmenting topologies (NEAT) \cite{stanley2002neat}. Our key results in this domain showcase that NSS-LC is far more robust and efficient than any other QD algorithm it is tested against, as it is able to find more solutions in more mazes and in fewer evaluations. Our findings suggest that introducing surprise to QD algorithms creates a new type of search that is more efficient, at least in the domain tested. The capacity of the NSS-LC algorithm seems to be due to the ability of surprise search to back-track, revisiting parts of the search space. Such a characteristic, combined with NS-LC, seems to yield a better balance between seeking greedily for entirely novel solutions and seeking for highly fit solutions locally.
	
\subsection{Novelty of this Paper}
This paper is novel in a number of ways. First, this is the first time that surprise search is used within a quality diversity algorithm. Second, the paper introduces three new QD algorithms with local competition, replacing novelty search with surprise search, or combining the two via aggregation or multiobjectivation. Third, the introduced algorithms---and all other baseline algorithms examined---are evaluated and compared comprehensively across 60 procedurally generated mazes of varying complexity and degrees of deceptiveness; this offers a broad assessment of their capacity, efficiency and robustness in this domain. Importantly, by procedurally generating the environments in which evolutionary (reinforcement) learning is tested we ultimately evaluate the generality of the algorithms' performance within the domain considered \cite{yannakakis2018artificial}. Earlier studies with novelty search \cite{lehman2011abandoning,mouret2011novelty}, surprise search \cite{gravina2016surprisebeyond,gravina2017noveltysurprise} and NS-LC \cite{pugh2016quality} in maze navigation focus merely on a few (up to 4) ad-hoc designed mazes. 
By necessity, the few mazes examined in such work represent certain aspects of the domain, limiting the generalizability of the findings. 
To the best of our knowledge, the only study that goes beyond a predefined set of mazes in the one by Lehman et al. \cite{lehman2011novelty} in which novelty search is compared against objective search in 360 maze-like environments; that early study however does not consider other forms of search (and QD algorithms in particular) for purposes of comparison. Fourth, no earlier study in divergent search and QD research has evaluated algorithms to the extensive degree we do in this paper. Our comparative analysis is comprehensive in \emph{depth}---i.e., each algorithm is evaluated and compared across a minimum of 3000 evolutionary runs in total---and in \emph{breadth}---i.e., our three introduced algorithms are compared against 6 other benchmark algorithms. The only other example is \cite{lehman2011novelty} which performed a total of 3600 evolutionary trials per algorithm, but only compared novelty search with objective search and random search; no other study has compared divergent search and QD algorithms to this degree. Finally, to the best of our knowledge this is the first study in which several QD algorithms are compared extensively for both their efficiency and robustness in maze navigation \cite{pugh2016quality, cully2018quality}, while a number of algorithmic variants are also tested to determine the impact of different parameters and components of the proposed QD algorithms. 

In summary, the core contribution of this paper is in the introduction of surprise as a mechanism for quality diversity in the form of three different algorithms, out of which one outperforms the state of the art. While the generality and extent to which we evaluate the algorithms are not an innovation per se they consist a decisive step towards establishing a methodology for evaluating QD algorithms based on procedurally generated content as well as providing an open-source testbed for QD algorithms. The large number of runs and the extensive comparisons with baselines and variations of the algorithms proposed similarly enhance the validity and generality of our findings. Further, the modular way algorithms are compared in this paper and the measures they are compared against (i.e., efficiency and robustness) collectively offer a holistic perspective on the algorithmic aspects that are responsible for any positive performance gain.  

\section{Related Work}\label{sec:background}
	
This section reviews work on divergence and quality diversity. Section \ref{sec:background_divergence} discusses why divergent search has been introduced for handling deceptive problems, while Section \ref{sec:background_qualitydiversity} describes the notion of quality diversity and the challenge of finding diverse and high-performing solutions.
	
\subsection{Divergence}\label{sec:background_divergence}
	
Divergence is a recent alternative in the context of EC that shifts from rewarding individuals fittest in terms of the objective (convergence) to pushing for diversity (divergence). As exploration of the search space progresses, rewarding diversity may lead to the right stepping stones towards the optimal solution without being attracted by the presence of suboptimal solutions (e.g., premature convergence to local optima). The problem of a greedy hill-climbing approach towards one objective that leads search away from a global optimum is neither new nor unique to evolutionary computation. However, the notion of \emph{deception} in the context of EC was introduced by Goldberg \cite{goldberg1987deceptive} to describe instances where highly-fit building blocks, when recombined, may guide search away from the global optimum. Similarly, the global optimum may be ``surrounded by'' (or composed of) genetic information that is deemed unfit. While this unfit information can be a vital stepping stone towards attaining the global optimum, it nevertheless may be ignored in favor of solutions which are fitter in the short term. 

Several solutions have been proposed in the literature to counter deception, from diversity maintenance techniques such as niching \cite{wessing2013niching} and speciation \cite{stanley2002neat} to multi-objective solutions. In multimodal optimization, for instance, the niching algorithm NEA2 \cite{preuss2015multimodal} uses nearest-better clustering to spread several parallel populations over the multiple local optima that are present in the fitness landscape. However, in \cite{lehman2013effective}, it is argued that in perversely deceptive problems, such as maze navigation and biped robot locomotion, genotypic diversity is not sufficient, as all the possible ``good" innovation steps towards the global optimum are punished by the optimization process. Divergent search tackles the problem directly by rewarding diversity at the phenotypical level, and a shift of paradigm is proposed where rewarding \emph{behavioral} diversity becomes the predominant driver of evolution.
	
It is argued that divergence, therefore, can tackle the problem of a deceptive fitness landscape or a deceptive fitness characterization by awarding diverse behaviors \cite{lehman2011abandoning} which may eventually lead to optimal results.  As a popular example of divergent search methods, \emph{novelty search} \cite{lehman2011abandoning} is inspired by open-ended evolution and rewards behaviors that have not been seen previously during search. Inspired by the notion of self-surprise for unconventional discovery, \emph{surprise search} \cite{yannakakis2016searching} instead rewards unexpected behaviors based on past behavioral trends observed during search. As discussed in Section \ref{sec:introduction}, novelty and surprise are conceptually orthogonal: the former looks to the past to assess whether a solution is actually new while the latter looks to the (near) past to find trends that are expected to carry on in the (near) future and assesses whether a solution breaks these expectations. Novelty search and surprise search as EC methods are also shown to explore the search space in different ways \cite{gravina2016surprisebeyond} and find different solutions (in terms of both behavior and genotype) to the same problem. The orthogonal nature of surprise and novelty is evidenced by the fact that combining the two in a search algorithm results in improved robustness and efficiency compared to novelty search or surprise search on their own in deceptive tasks \cite{gravina2017noveltysurprise}.
	
\subsection{Quality Diversity}\label{sec:background_qualitydiversity}
    
While diversity alone can be beneficial to discover the optimal solution of a deceptive problem, for particular search spaces this is not enough. When the search space is boundless or when the diversity measure is completely uncorrelated to desired behaviors, some push towards quality is needed. Several solutions have been proposed to solve this issue, such as Novelty Multiobjectivation \cite{mouret2011novelty} or constrained novelty search \cite{lehman2010revising,liapis2015ecj}. However, a more aligned combination of divergence and convergence might enable evolutionary search to discover high-performing and diverse solutions in the same run. Inspiration for handling this problem is again found in nature. Natural evolution has discovered an impressive number of diverse solutions (i.e., organisms) that can adapt to different conditions and constraints. If we take for example the problem of moving, nature has evolved different ways to ambulate: crawling, swimming, walking, etc. Moreover, an organism can be composed of different legs. It becomes apparent that each configuration of the above can lead to an effective yet different solution to movement.
	
Inspired by the richness of solutions found in nature, Pugh et al. \cite{pugh2015confronting, pugh2016quality} have proposed the quality diversity challenge. The main purpose of quality diversity is the discovery of \emph{multiple and diverse good solutions} at the same time. Notable examples of QD algorithms include novelty search with local competition \cite{lehman2011creatures} and MAP-Elites \cite{mouret2015illuminating} which explore ways to combine the search for quality (within the locality of the solution's space) with the search for the diversity of solutions.
	
The quality diversity paradigm has been inspired by the idea of solving single-objective optimization problems via multi-objective optimization (MOO). Such approaches have been successfully applied to different problems in the literature. For instance, \cite{gong2015multiobjective} proposes using MOO to optimize two conflicting objectives, the reconstruction error and the sparsity of deep artificial neural networks. In \cite{li2016multi}, \emph{multi-objective self-paced learning} decomposes a hard problem into simpler problems which can be optimized in a more accessible way. In \cite{qian2015pareto}, a multi-objective approach is proposed for ensemble learning to obtain the best performance with the fewest learners. In \cite{qian2015subset}, feature selection is performed through evolutionary Pareto optimization to select the best possible subset of features for a deep neural network (feature learning). Finally, in \cite{qian2017constrained} MOO is used for influence maximization in a constrained scenario. We argue that a QD algorithm will discover more highly-fit solutions if we extend the divergence pressure across multiple and orthogonal dimensions beyond novelty---such as surprise. Following this view, we introduce three algorithms that explore how both surprise and a combination of novelty and surprise might help to cover the solution space more advantageously.
	
\section{Algorithms}\label{sec:algorithms}
	
This section reviews work on EC algorithms targeting divergence and quality diversity, and introduces three novel approaches in the quality diversity field.
	
\subsection{Background Algorithms}\label{sec:algorithms_background}

This section discusses algorithms used for comparative purposes and algorithms which are enhanced in this paper. We first present two implementations of divergent search: novelty search and surprise search. Finally, we describe an established QD algorithm, novelty search with local competition.
	
\subsubsection{Novelty Search}\label{sec:algorithms_divergence_novelty}
	
Novelty search \cite{lehman2011abandoning} is the first divergent search algorithm that has been proposed with demonstrated effectiveness in domains such as maze navigation and robot control \cite{lehman2011abandoning}, image generation \cite{lehman2012impressiveness} and game content generation \cite{liapis2013delenox, liapis2015ecj}. Novelty search ignores the objective of the problem at hand, and attempts to maximize behavioral diversity in terms of a novelty score. The novelty score is computed through Eq.~\eqref{eq:novelty}, i.e., the average distance with the $n_{NS}$ closest individuals in the current population or an archive of novel solutions. In every generation, individuals with a novelty score above a fluctuating threshold (parameter $\rho$) are added to a novelty archive which persists throughout the evolutionary run. The distance $d_n$, which is used to assess the novelty score as well as what constitutes an individual's closest neighbors, is based on the difference of behaviors (rather than genotypes) between individuals. This allows novelty search to explore a diverse set of behaviors without explicitly favoring behaviors closer to the desired behavior, i.e., the solution of a problem.
\begin{equation}
n(i) = \frac{1}{n_{NS}}\sum_{j=0}^{n_{NS}}d_n(i, \mu_j)\label{eq:novelty}, 
\end{equation}
\noindent where $d_n$ is the behavioral distance between two individuals and depends on the domain under consideration (e.g., the Euclidean distance between two robots' final positions in a maze navigation task), $\mu_j$ is the j-$th$ nearest neighbor and $n(i)$ is the novelty score for the individual $i$. The current population and the novelty archive are used to find the nearest neighbors.
\subsubsection{Surprise Search}\label{sec:algorithms_divergence_surprise}

Surprise search is a recent divergent search algorithm \cite{yannakakis2016searching} which has demonstrated its effectiveness and robustness in domains as diverse as maze navigation \cite{gravina2016surprisebeyond, gravina2017surprise}, robot morphology evolution \cite{gravina2017soft} and procedural content generation \cite{gravina2016constrained}. Surprise search is built on the novelty search paradigm, but instead of deviating from past and current behaviors (in the novelty archive or the current population, respectively) it attempts to deviate from predicted behaviors based on recent trends. Inspired by the way humans experience surprise, the algorithm operates on the definition of surprise as ``deviation from the expected''. Unlike novelty, surprise hinges on a sequence of patterns and their temporal order: the same artifact such as a green circle could be perceived as non-surprising if the last produced shape was green and curvy, but it would be surprising if past produced shapes alternated between blue circles and green squares \cite{yannakakis2016searching}.
	
Surprise search requires two components: a prediction model and a deviation model. The prediction model uses past behaviors to estimate future behaviors. More specifically, the prediction model is derived from a selected number of past generations $g$ (parameter $h$ as history length) and their locality in the behavioral space (behavior vector $\textbf{b}_g$ of size $k_{SS}$). Through this model $m$ summarized in Eq.~\eqref{eq:prediction_model}, a number of predictions (vector $\mathbf{p}$ of size $k_{SS}$) are made based on past trends for the expected behaviors in the current population. Once predictions are made, the deviation model rewards each individual in the current population based on its distance from the $n_{SS}$ closest predictions as per Eq.~\eqref{eq:surprise}. As with novelty search, the distance $d_s$ used in the deviation model is based on the difference in behaviors; similarly, the prediction model creates expectations in terms of behavior. This two-step mechanism favors individuals that diverge from predicted future trends; it should be noted that based on $m$, $h$ and $k_{SS}$ parameters, the predictions could be infeasible or unreachable by conventional search (e.g., points outside a maze that should be traversed). More details on surprise search can be found in \cite{yannakakis2016searching, gravina2016surprisebeyond, gravina2017surprise}.
\begin{equation}
\mathbf{p} = m(h, k_{SS}) \label{eq:prediction_model}
\end{equation} 
\begin{equation}
s(i)  = \frac{1}{n_{SS}}\sum_{j=0}^{n_{SS}}d_s(i,p_{i,j})\label{eq:surprise}, 
\end{equation} 
\noindent where $s(i)$ is the surprise score of individual $i$, and $d_s$ is the behavioral distance of $i$ from its $n$ closest predictions ($p_{i,j}$).

\begin{figure*}[!tb]
\centering
\includegraphics[width=1\textwidth]{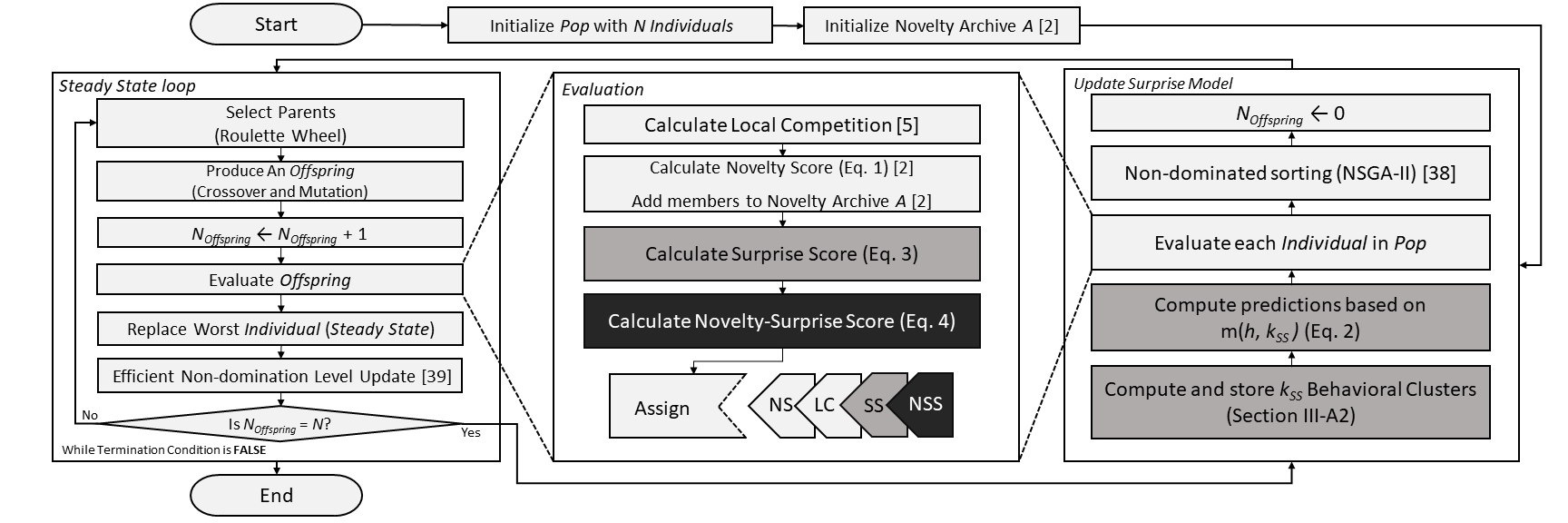}
\caption{\textbf{Surprise-based Quality Diversity}. The flow chart illustrates a high level representation of the three introduced QD algorithms: SS-LC, NSS-LC, NS-SS-LC. All algorithms are based on a steady state evolutionary algorithm framework (left flow chart). The model of surprise is initialized after the generation of the initial population and then updated every $N$ offspring generations (right flow chart). The evaluation of individuals (middle flow chart) goes through the calculation of local competition, novelty, and surprise scores before those are assigned to the corresponding algorithm.
{The introduced surprise-based components and novelty-surprise components of the algorithm are depicted in grey boxes and in black boxes respectively.}
References are made to parts of existing divergent search and QD algorithms, as well as corresponding equations and sections in this paper.
}
\label{fig:qd_steady_state}
\end{figure*}
	
\subsubsection{Novelty Search with Local Competition}\label{sec:algorithms_qualitydiversity_nslc}

Considered to be the first QD algorithm, novelty search with local competition \cite{lehman2011creatures} is an algorithm that combines the divergence of novelty search with the localized convergence obtained through a local competition. In NS-LC, a multi-objective algorithm, NSGA-II \cite{deb2002fast}, searches for non-dominated solutions across two dimensions: novelty and local competition. Novelty attempts to maximize a novelty score computed in Eq.~\eqref{eq:novelty}, i.e., the average distance of the individual's behavior with the behavior of the closest neighbors in the current generation or the novelty archive. Local competition is also calculated based on the closest individuals in the current generation and the novelty archive. The reward for local competition $lc(i)$ is proportional to the number of solutions outperformed in terms of the objective function $f$. This creates a pressure towards those solutions that are good within their (behavioral) niche, even if they globally underperform compared to the general population. Combining novelty search and local competition allows NS-LC to pursue and optimize many different behaviors in the hope that one of these directions will eventually lead to globally optimal solutions. NS-LC has shown performance advantages over novelty search in the domains of maze navigation \cite{pugh2015confronting, pugh2016quality} and robot evolution \cite{lehman2011creatures, cully2018quality}.
	
\subsection{Surprise for Quality Diversity}\label{sec:algorithms_QD_surprise}
	
Quality diversity algorithms are designed to couple divergence with pressure for a local measure of quality. Inspired by previous work on divergent search for surprise \cite{gravina2017surprise, gravina2017noveltysurprise}, we propose three new algorithms that introduce surprise search as an alternative divergent search mechanism with local competition for quality diversity \cite{lehman2011creatures}. These three algorithms are named surprise search with local competition (SS-LC), novelty-surprise search with local competition (NSS-LC), and the three-objective novelty search-surprise search-local competition (NS-SS-LC). 
    
Fig.~\ref{fig:qd_steady_state} offers a high-level overview of the three algorithms used for surprise-based QD. The algorithms initially generate a population ($Pop$) of $N$ individuals and then initialize a novelty archive ($A$). Then an initial update of the surprise model is performed (right part of Fig.~\ref{fig:qd_steady_state}; this phase is described in detail below) and each individual in the population is evaluated according to the selected algorithm (i.e., NSS-LC, SS-LC or NS-SS-LC). Right before entering the main loop of the algorithm, non-dominated sorting divides the population into fronts based on the NSGA-II algorithm. While the termination criterion is not met (e.g., high performance is reached) a steady-state evolutionary algorithm is executed (see left part of Fig. \ref{fig:qd_steady_state}): two mating parents are selected to generate a new offspring, which is evaluated based on the algorithm chosen (i.e., NSS-LC, SS-LC or NS-SS-LC); the worst individual of the population is then replaced by the newly generated offspring if the offspring is more fit. At the end of each steady state step of the algorithm, the non-dominated fronts are updated as in \cite{li2017efficient}. Every $N$ offspring generations, we update the surprise model (see right part of Fig.~\ref{fig:qd_steady_state}) by computing the $k_{SS}$ behavioral clusters and then computing the predictions based on the surprise predictive model $m$. We then re-evaluate the entire population and recompute the non-dominated fronts. The algorithm returns to the main steady state loop until either $N$ new offspring are generated or the termination condition is reached. A key part of the framework is its evaluation step (see middle flow chart in Fig.~\ref{fig:qd_steady_state}): it involves the computation of novelty, surprise, novelty-surprise and local competition scores. Based on the algorithm chosen, we assign the selected scores to the individual which will influence the replacement strategy and the non-dominated sorting performed by the NSGA-II algorithm. The details of each of the three algorithms are described in the subsections below, whereas their domain-specific implementation is further elaborated in Section \ref{sec:maze_domain_algorithms}.

\subsubsection{Surprise Search with Local Competition}\label{sec:algorithm_sslc}
	
As a direct integration of surprise search for quality diversity, the obvious approach is to replace novelty with surprise in the NS-LC \cite{lehman2011creatures} paradigm. In surprise search with local competition (SS-LC), NSGA-II \cite{deb2002fast} searches for non-dominated solutions on the dimensions of local competition and surprise. Local competition is calculated based on the superiority of the individual being evaluated among its closest neighbors. What constitutes a nearby neighbor for local competition is based on a behavioral characterization ($d_{SS}$ in this case), rather than a genotypic one. Superiority is established based on a measure of proximity to an ideal solution: the number of neighbors who are worse than the current individual is used as the local competition score to be optimized. Borrowing from local competition as applied in NS-LC, the closest neighbors are drawn from the current population. Unlike NS-LC, however, SS-LC does not maintain a novelty archive and only considers neighbors in the current population. The surprise dimension uses the surprise score of eq.~\eqref{eq:surprise}, which assesses how much the behavior of an individual deviates from expected behaviors based on trends in recent generations. As described in Section \ref{sec:algorithms_divergence_surprise}, the surprise score is calculated based on a two-step process: first, Eq.~\eqref{eq:prediction_model} creates predictions based on past generations' dominant behaviors, and then Eq.~\eqref{eq:surprise} calculates the surprise score based on the distance from predicted behaviors.
	
\subsubsection{Novelty-Surprise Search with Local Competition}\label{sec:algorithm_nsslc}
	
Given the orthogonal nature of novelty and surprise, we hypothesize that combining novelty and surprise as different measures of divergence would be valuable for QD. Earlier work has shown that a weighted sum of the surprise score and the novelty score can outperform novelty search alone \cite{gravina2017noveltysurprise}. We thus expect that combining both surprise and novelty as measures of divergence with local competition can only improve the performance of the state of the art QD algorithm.
	
In the novelty-surprise search with local competition (NSS-LC) algorithm, as we name it, NSGA-II \cite{deb2002fast} searches for non-dominated solutions on the dimensions of local competition and a weighted sum combining novelty and surprise. Local competition measures the number of closest neighbors (in terms of behavior) which underperform compared to the current individual. Unlike SS-LC, local competition considers the novelty archive maintained by the novelty search component and thus the closest neighbors considered for local competition can be from both the current population and the novelty archive. The other dimension targeted by NSGA-II combines the novelty score of Eq.~\eqref{eq:novelty} and the surprise score of Eq.~\eqref{eq:surprise} in the weighted sum of Eq.~\eqref{eq:novelty_surprise}, where a single parameter ($\lambda$) influences both scores.
\begin{equation}
ns(i) = \lambda \cdot n(i) + (1 - \lambda) \cdot s(i) \label{eq:novelty_surprise},
\end{equation}
\noindent where $ns(i)$ is the combined novelty and surprise score of individual $i$ and $\lambda \in [0, 1]$ is a parameter that controls the relative importance of novelty versus surprise, $n(i)$ is the novelty score (Eq. \ref{eq:novelty}) and $s(i)$ is the surprise score (Eq. \ref{eq:surprise}).

\subsubsection{Novelty Search--Surprise Search--Local Competition}\label{sec:algorithm_nssslc}

As an alternative way to combine novelty search and surprise search in the divergence dimension of a QD algorithm, we can consider them as independent objectives rather than aggregating them as with NSS-LC. The three-objective algorithm Novelty Search--Surprise Search--Local Competition (NS-SS-LC) uses NSGA-II \cite{deb2002fast} to search for non-dominated solutions on the three dimensions: the local competition score, the surprise score (as in Eq.~\ref{eq:surprise}) and the novelty score (as in Eq.~\ref{eq:novelty}). 
	
\section{Maze Navigation Testbed}\label{sec:maze_domain}
	
One of the most popular testbeds for divergent search and quality diversity algorithms is the maze navigation problem. First proposed in \cite{lehman2011abandoning}, the maze navigation problem has several properties that make it suitable for testing QD algorithms and, hence, our hypothesis. This section first describes the robotic controller which performs the maze navigation task, along with details of the neuroevolutionary approaches for evolving the controller. The remainder of this section describes an algorithm for generating deceptive mazes for the controller to navigate, the criteria for deceptiveness in selecting the mazes and the final set of mazes on which all algorithms are tested.
	
\subsection{Robot Controller}\label{sec:maze_domain_navigation}
	
The problem of maze navigation can be simply formulated as follows: a robot starting at a specific position in the maze must reach the goal position in a maze using local (incomplete) information. The robot is equipped with six range finder sensors which indicate the distance from the closest wall and four pie-slice sensors broadly indicating the direction of the goal. During simulation, the sensors' data is provided as input to an artificial neural network (ANN) which controls the movement of the robot, i.e., its velocity and turning angle as two outputs. As per the problem formulation, the \emph{de facto} objective is to reach the goal position, and the most intuitive objective/fitness function is to select individuals based on their Euclidean distance from the goal---ignoring the presence of walls. The fact that the maze topology is not known in advance (local information) results in a deceptive fitness landscape as the robot may end up in a dead-end which is locally optimal but is unable to bring the robot any closer to the goal. In order to find the goal, in most cases the robot must go through areas of lower fitness before the goal becomes accessible. Deceptive mazes are easy to identify visually, as they feature dead ends along the direct line between starting and goal position. However, deceptive mazes are challenging to identify computationally; we discuss this in Section \ref{sec:maze_domain_generation}. Beyond a visually interpretable search space, which largely coincides with the physical space of the maze, an additional property of this testbed is the relatively lightweight simulations it affords; this allows extensive tests to be run, featuring evolutionary runs with large population sizes and multiple re-runs. Indicatively, this paper performs 50 independent evolutionary runs per maze (or more, considering a sensitivity analysis), and tests 60 mazes in this fashion. Such a computational burden is prohibiting in more complex simulations such as evolving robot morphologies \cite{lehman2011creatures,gravina2017soft} or generating game content and testing it in games \cite{gravina2016constrained, cardamone2011evolving}.
	
As with the original implementation of the maze navigation testbed \cite{lehman2011abandoning}, the ANN of the robot controller is optimized through neuroevolution of augmenting topologies \cite{stanley2002neat} (NEAT). NEAT starts with an initial population of simple networks and progressively increases their complexity by adding nodes and edges. NEAT ensures genotypic diversity through speciation, which groups networks with similar topologies into species. In this paper, genetic operators, mutation chances and speciation parameters are identical to those reported in \cite{lehman2011abandoning}; parameters of novelty search are described in Section \ref{sec:experiments}.
	
\subsection{Algorithms for Maze Navigation}\label{sec:maze_domain_algorithms}
	
This section describes the implementation details of the algorithms which are compared in Section \ref{sec:experiments} to test our hypothesis that surprise search enriches the capacity of QD. The source code of the described framework is available online\footnote{https://gitlab.com/2factor/QDSurprise}. It is based on the source code of the original novelty search implementation for maze navigation \cite{lehman2011abandoning} and implements all the algorithms described in this work.
		
\subsubsection{Objective Search}
	
As an indication of how deceptive landscapes can hinder traditional EC approaches, objective search is used as a baseline for selecting among generated mazes in Section \ref{sec:maze_domain_generation}. As per \cite{lehman2011abandoning}, the objective $f$ is to minimize the Euclidean distance between  the goal in the maze and the robot's final position at the end of simulation. Objective search never finds a solution in any of the mazes used in the experiments of Section \ref{sec:experiments}, so its results are omitted.
	
\subsubsection{Novelty Search}
	
As discussed in Section \ref{sec:algorithms_divergence_novelty}, novelty search ignores the objective of the problem at hand and attempts to maximize behavioral diversity. The novelty score, computed through Eq.~\eqref{eq:novelty}, uses the behavioral distance $d_n$ between two individuals; in this domain $d_n$ is the Euclidean distance between the two robots' final positions at the end of simulation. The same distance is used to identify closest neighbors ($\mu_j$). Finally, in all novelty search experiments in this paper we follow the literature and consider the $15$ closest individuals as our $n_{NS}$ parameter \cite{lehman2011creatures}.
	
\subsubsection{Novelty Search with Local Competition}
	
As discussed in Section \ref{sec:algorithms_qualitydiversity_nslc}, novelty search with local competition combines the divergence of novelty search with the localized convergence obtained through a local competition \cite{lehman2011creatures}. In this paper, NS-LC uses a steady-state NSGA-II multi-objective algorithm \cite{li2017efficient} to find non-dominated solutions on two dimensions: novelty and local competition. Novelty attempts to maximize a novelty score computed in Eq.~\eqref{eq:novelty}, using the Euclidean distance between the two robots' final positions at the end of simulation for calculating distance from the closest individuals. As with novelty search, we are based on the successful experiments performed in the literature \cite{lehman2011creatures} and consider the $n_{NS}=15$ closest individuals in all NS-LC experiments of this paper. Local competition is calculated based on the closest individuals (in terms of Euclidean distance) in the current population and the novelty archive. In the maze navigation task, local competition counts the number of neighboring robots with final positions that are farther from the goal in terms of Euclidean distance. Note that unlike earlier work, this paper decouples the number of individuals considered for the novelty score ($n_{NS}$) with the number of individuals used to calculate local competition ($n_{LC}$). Section \ref{sec:experiments} performs a sensitivity analysis for the best locality parameter ($n_{LC}$) of local competition in NS-LC and the other QD algorithms.
	
\subsubsection{Surprise Search with Local Competition}
	
\begin{figure}[tb]
	\centering
	\subfloat[Generation $g - 2$]{
	\includegraphics[width=0.31\linewidth]{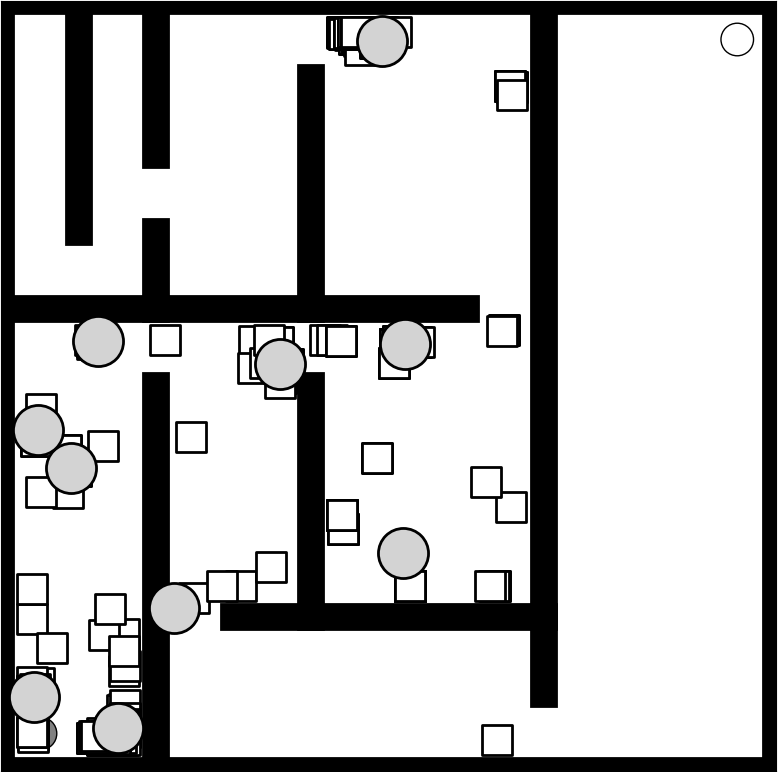}}
	\subfloat[Generation $g - 1$]{
	\includegraphics[width=0.31\linewidth]{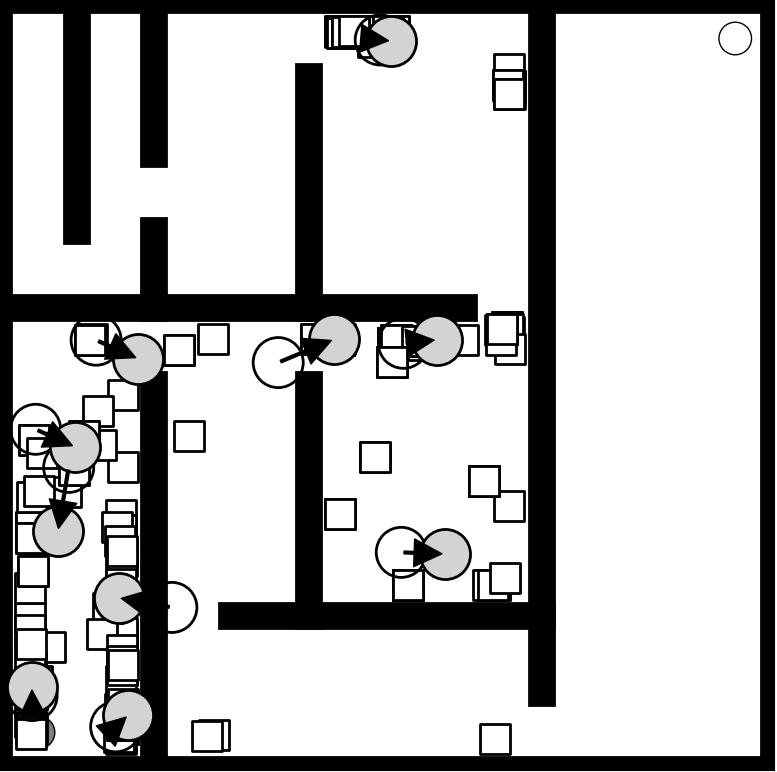}}
    \subfloat[Generation $g$]{
    \includegraphics[width=0.31\linewidth]{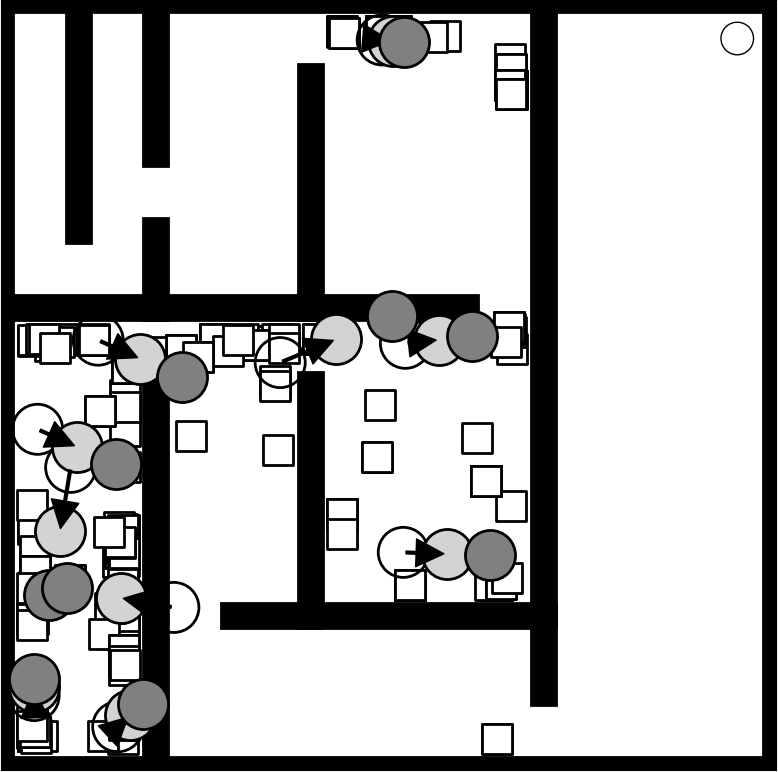}}
    \caption{The key phases of the surprise search algorithm as applied to the maze navigation domain. Surprise search uses a history of two generations ($h=2$) and 10 behavioral clusters ($k_{SS}=10$) in this example. Robots' final positions are depicted as empty squares; cluster centroids and prediction points are depicted as light grey and dark grey circles, respectively.}
    \label{fig:surprise_search}
\end{figure}

As noted in Section \ref{sec:algorithm_sslc}, surprise search with local competition (SS-LC) uses a steady-state NSGA-II multi-objective algorithm \cite{li2017efficient} to find non-dominated solutions on two dimensions: surprise and local competition. Surprise uses the prediction model of Eq.~\eqref{eq:prediction_model} to make a number of predictions ($\textbf{p}$) for the expected behaviors in the current population. The individuals in the current population are then evaluated based on their distance from the $n$ closest predictions as per Eq.~\eqref{eq:surprise}. In this testbed, predictions are made on the robots' final position, and distance refers to the Euclidean distance between two robots' final positions at the end of simulation. For deriving $k_{SS}$ behaviors to predict, k-means clustering is applied on the robots' final position in one generation. Based on earlier research on surprise search and novelty-surprise search on difficult maze navigation tasks \cite{gravina2017noveltysurprise}, $k_{SS}=200$ and $n_{SS}=2$ in this paper. The two last generations are used ($h=2$) to cluster behaviors, and predictions are based on a linear interpolation between cluster centroids in subsequent generations (Fig.~\ref{fig:surprise_search} illustrates how behaviors are clustered and how predictions are computed). More details on the way clustering is performed and predictions are made for surprise search in maze navigation can be found in \cite{gravina2017surprise, gravina2016surprisebeyond}.

	\begin{table}[!tb]
	
	\centering
	
	\caption{Ten easiest generated mazes sorted by NS-LC successes. 
    The starting position (grey filled circle) is at the bottom left corner; the goal position (black empty circle) is at the top right corner. The three numbers under each maze from left to right represent, respectively, the number of subdivisions, the A* path length, and the number of successes of NS-LC.}
	\label{fig:maze_example}
	
	\bgroup
	
	\newcommand \size{0.155\linewidth}
   \newcommand\maze{1}
	
	\begin{tabular}{c c c c c}
    \parbox[c][\size]{\size}{\includegraphics[width=\maze\linewidth]{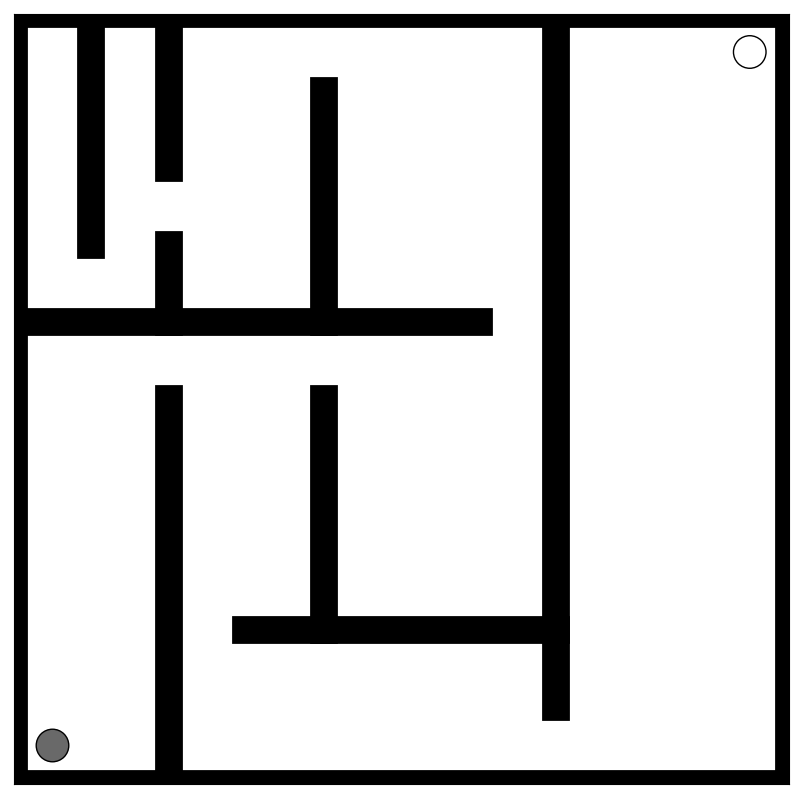}} &
    \parbox[c][\size]{\size}{\includegraphics[width=\maze\linewidth]{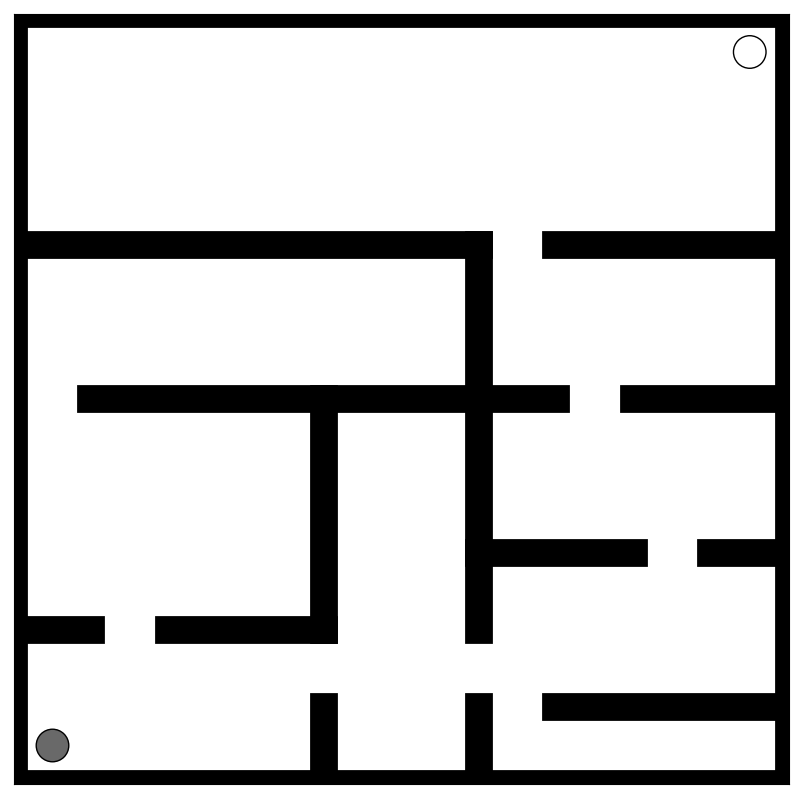}} &
    \parbox[c][\size]{\size}{\includegraphics[width=\maze\linewidth]{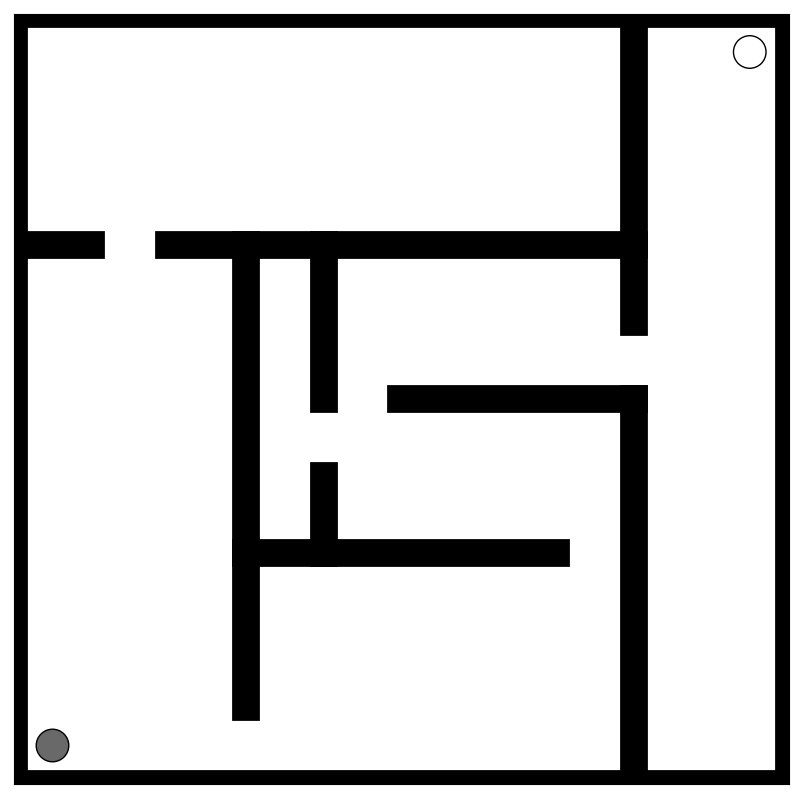}} &
    \parbox[c][\size]{\size}{\includegraphics[width=\maze\linewidth]{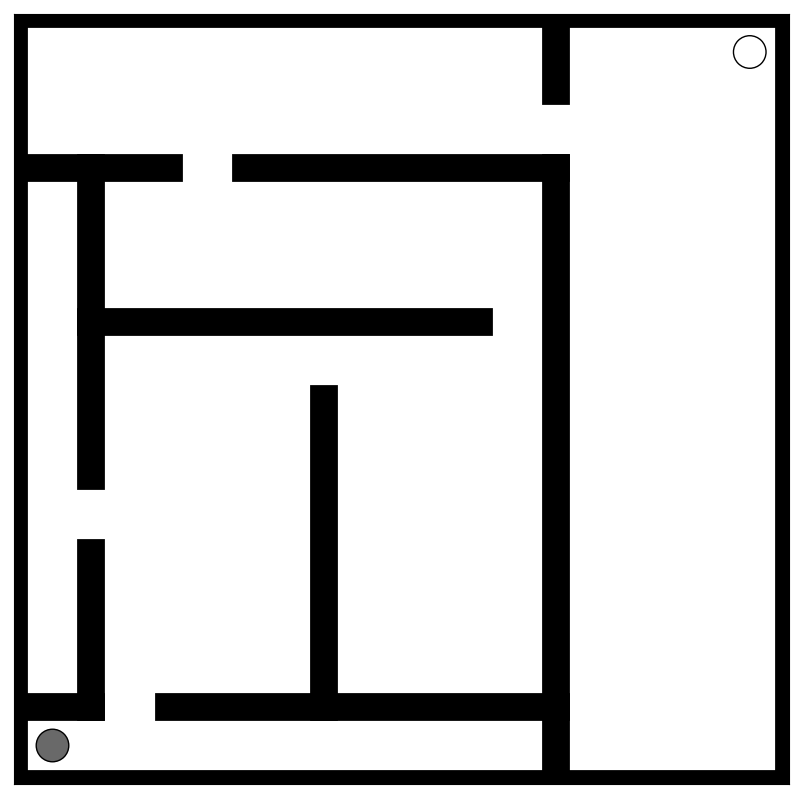}} &
    \parbox[c][\size]{\size}{\includegraphics[width=\maze\linewidth]{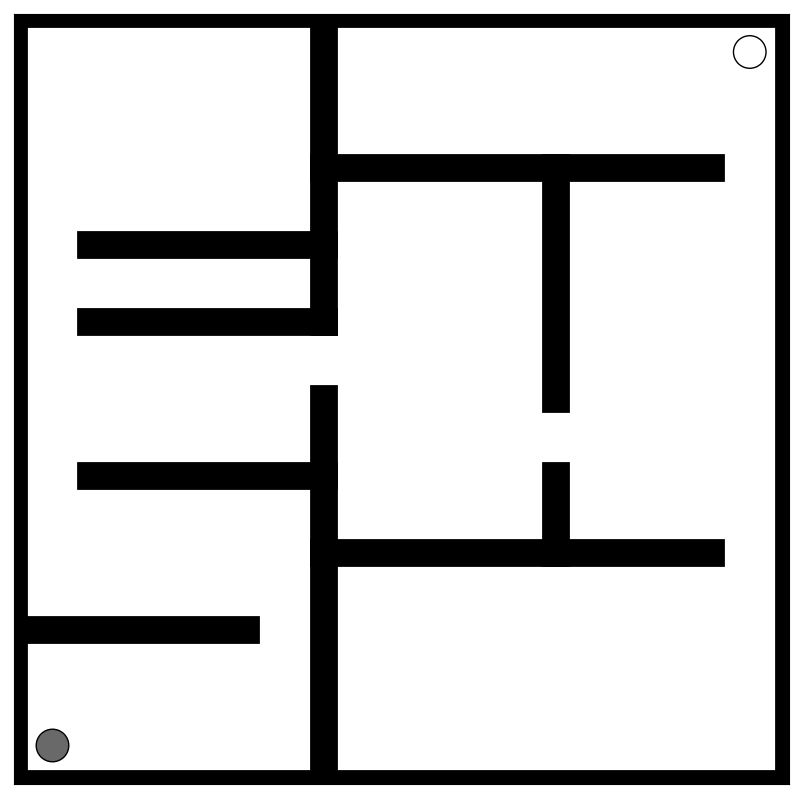}}\\
 $\left[8, 453, 50\right]$ & $\left[8, 332, 40\right]$ & $\left[6, 362, 32\right]$ & $\left[6, 391, 22\right]$ & $\left[8, 364, 17\right]$ \\ 
    \parbox[c][\size]{\size}{\includegraphics[width=\maze\linewidth]{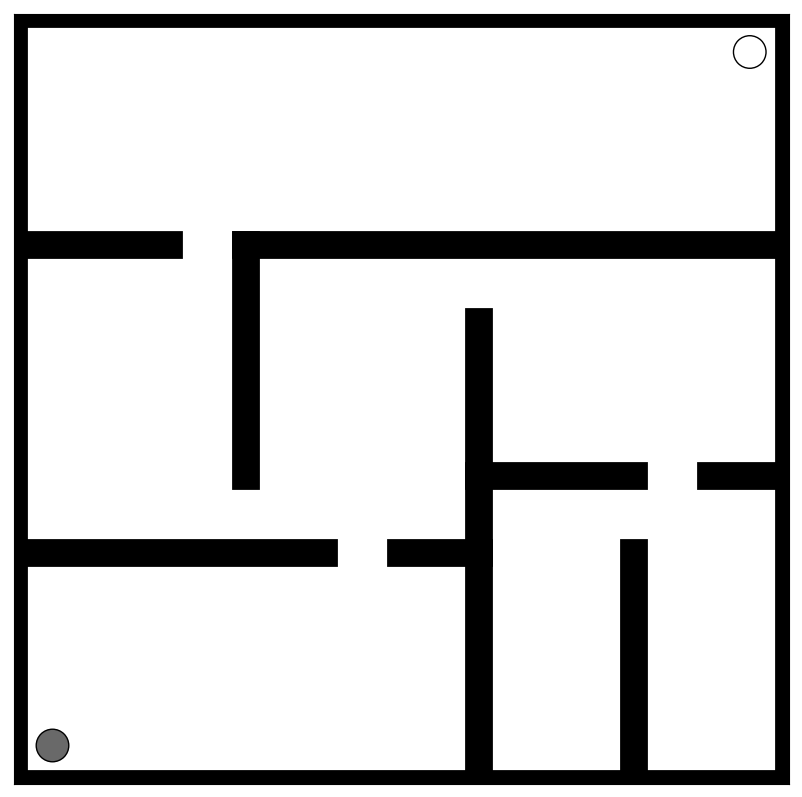}}  &
    \parbox[c][\size]{\size}{\includegraphics[width=\maze\linewidth]{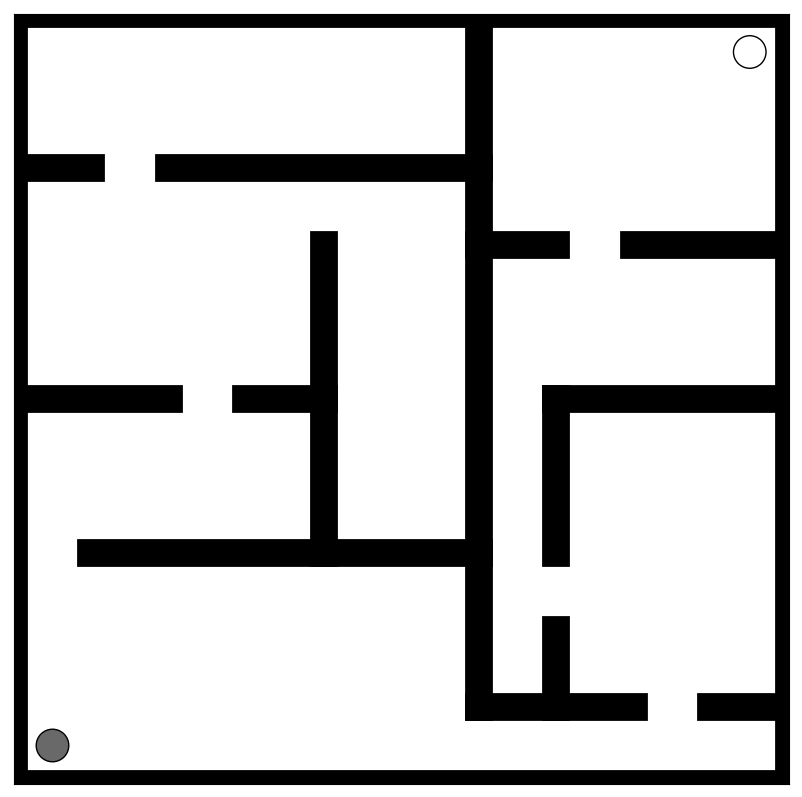}}  &
    \parbox[c][\size]{\size}{\includegraphics[width=\maze\linewidth]{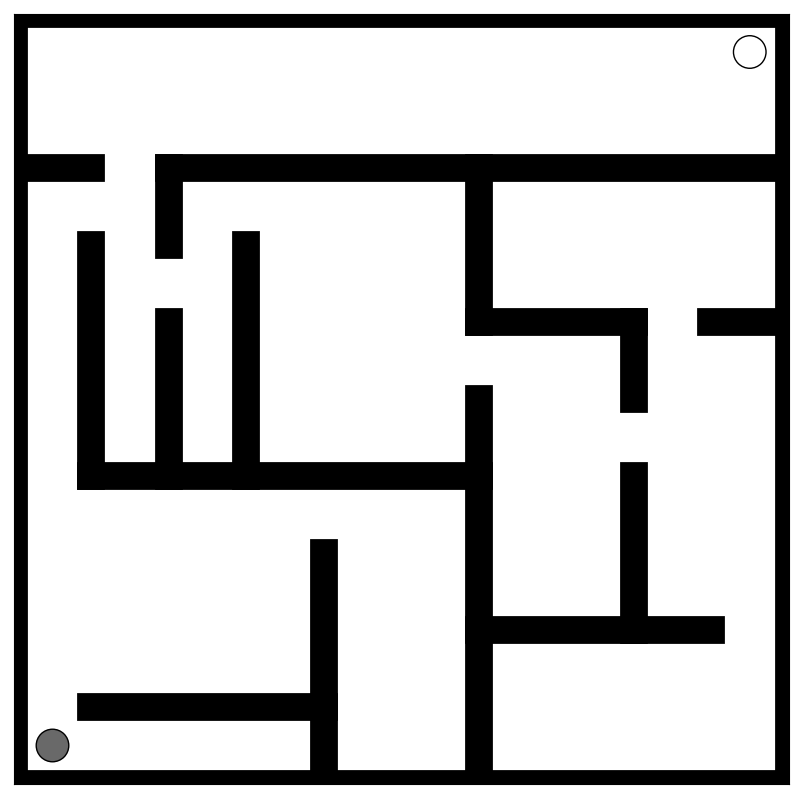}}  &
    \parbox[c][\size]{\size}{\includegraphics[width=\maze\linewidth]{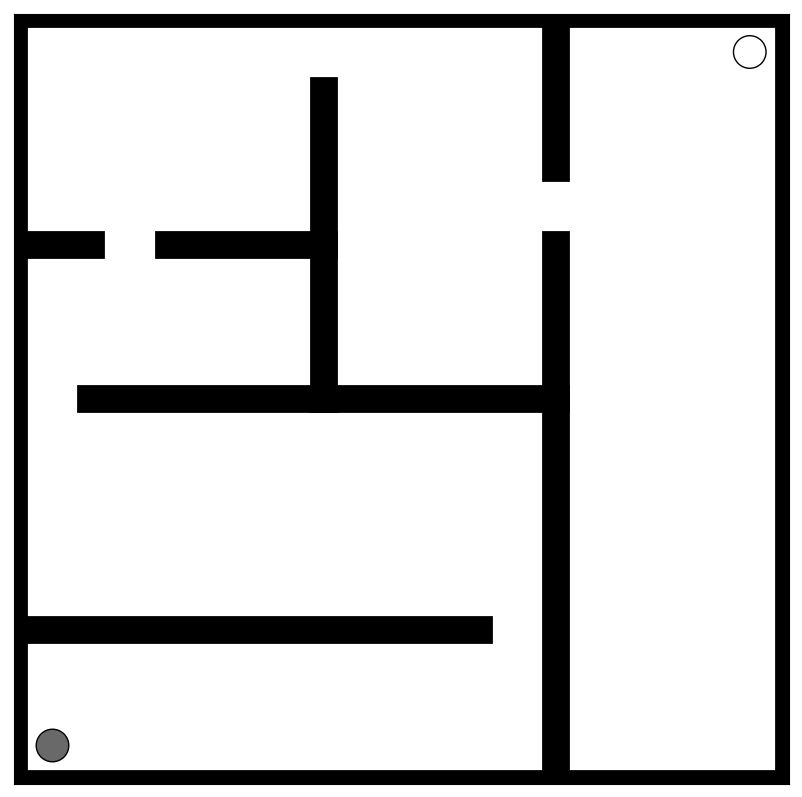}} &
    \parbox[c][\size]{\size}{\includegraphics[width=\maze\linewidth]{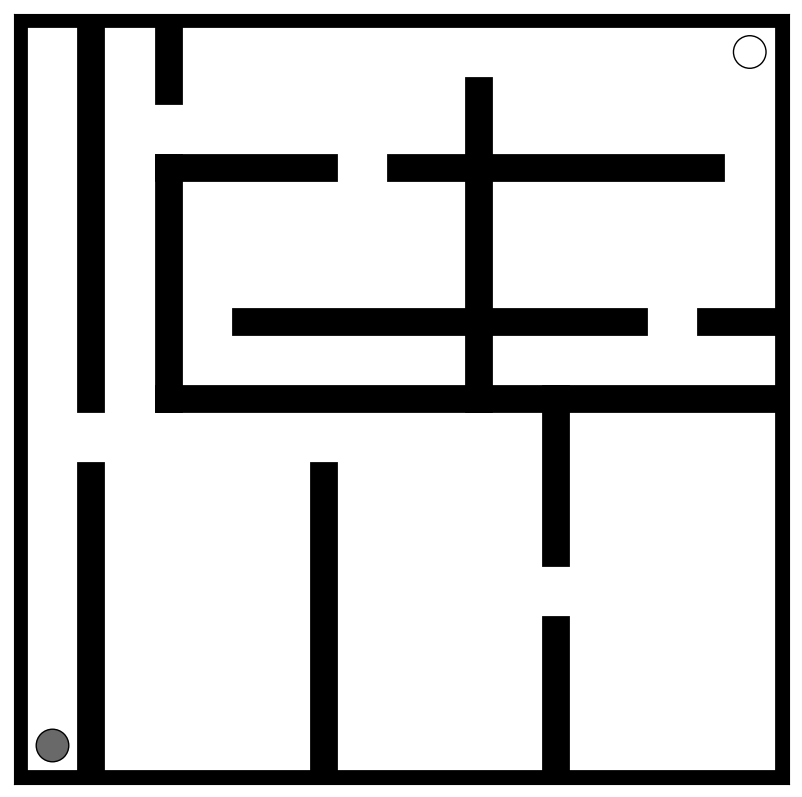}} \\
$\left[6, 333, 16\right]$ & $\left[9, 355, 15\right]$ & $\left[11, 325, 14\right]$ & $\left[5, 488, 12\right]$ & $\left[10, 325, 5\right]$ \\
    \end{tabular}
	\egroup
\end{table}
    
	\subsubsection{NSS-LC}
	As noted in Section \ref{sec:algorithm_nsslc}, novelty-surprise search with local competition (NSS-LC) uses a steady-state NSGA-II multi-objective algorithm \cite{li2017efficient} to search for non-dominated solutions on the dimensions of local competition (computed in the same way as NS-LC) and a weighted sum of the novelty score and surprise score as in Eq.~\eqref{eq:novelty_surprise}. Parameters for novelty search are the same as in \cite{lehman2011creatures} (i.e., $n_{NS}=15$), while parameters for surprise search are the same as above (i.e., $h=2$, $n_{SS}=2$, $k_{SS}=200$). Distance characterizations for NSS-LC are computed as in the other QD algorithms, via the Euclidean distance between the two robots' final positions at the end of simulation. Experiments in Section \ref{sec:experiments_sensitivity} identify which values for $\lambda$ and $n_{LC}$ (the neighbors considered for local competition) are most appropriate for this testbed.
	
	\subsubsection{NS-SS-LC} 
	As noted in Section \ref{sec:algorithm_nssslc}, NS-SS-LC uses a steady-state NSGA-II multi-objective algorithm \cite{li2017efficient} to search for non-dominated solutions on three dimensions: local competition (computed in the same way as NS-LC), novelty search, and surprise search. Parameters for novelty search are the same as in \cite{lehman2011creatures} (i.e., $n_{NS}=15$), while parameters for surprise search are the same as above (i.e., $h=2$, $n_{SS}=2$, $k_{SS}=200$). Distance characterizations for NS-SS-LC are computed as in the other QD algorithms, via the Euclidean distance between the two robots' final positions at the end of simulation. Experiments in Section \ref{sec:experiments_sensitivity} identify the best $n_{LC}$ values for NS-SS-LC in this testbed.
	
	\subsection{Maze Generation}\label{sec:maze_domain_generation}
	
	While a substantial portion of research on divergent search and quality diversity have focused on hand-crafted deceptive mazes \cite{lehman2011abandoning,lehman2010revising,pugh2016quality,gravina2016surprisebeyond,gravina2017noveltysurprise}, this paper uses a broader set of mazes to test the QD algorithms of Section \ref{sec:maze_domain_algorithms}. These mazes are not crafted by a human designer but generated procedurally; a similar set of generated mazes has been used in \cite{lehman2011novelty} to evaluate the performance of divergent search. Generating rather than hand-crafting mazes allows for a broader range of spatial arrangements to be tested, without human curation towards possibly favorable patterns. Additionally, this paper uses maze generation to find appropriate deceptive mazes which satisfy two criteria related to the algorithms tested rather than inherent structural patterns. These criteria are:
	\begin{enumerate}
		\item Objective search must not find any solutions in 50 evolutionary runs.
		\item NS-LC (with parameters as in \cite{lehman2011creatures}) must find a solution in at least one of 50 evolutionary runs.
	\end{enumerate}
	The first criterion establishes that each maze is deceptive, as attempting to optimize proximity to the goal does not result in any solutions. However, the first criterion may be satisfied by mazes which have extremely long paths from start to finish which may not be reachable within the alloted simulation time for each robot. To ensure that the maze is not too difficult (or practically impossible), the state-of-the-art QD algorithm NS-LC is used as a second criterion: if ``vanilla'' NS-LC \cite{lehman2011creatures} cannot find a solution even after numerous retries, then the maze is characterized as too difficult and is ignored.
	
	Mazes are generated via a recursive division algorithm \cite{reynolds2010maze}, which subdivides the maze (by adding walls at the border) recursively until no more walls can be added. In this paper, the algorithm stops after a specific number of subdivisions, or when adding a new wall would make the path non-traversable. The start position is always set on the bottom-left corner and the goal position on the top-right corner. Width of gaps and the minimum width of corridors is defined in a way that allows the robot controller to comfortably pass through. All mazes tested have between 5 and 12 subdivisions (chosen randomly), and evolution for both objective search and NS-LC is performed on a population of 250 individuals for a maximum of 600 generations and a simulation time of 300 frames.
    
\begin{table}[tb]
  \centering
  \caption{Distribution of the 60 selected mazes and corresponding average length of their shortest path, as computed by A*, across the number of subdivisions.}
  \begin{tabular}{l||@{ }c@{ }@{ }c@{ }@{ }c@{ }@{ }c@{ }@{ }c@{ }@{ }c@{ }@{ }c@{ }@{ }c@{ }}
  \hline\hline
  Subdivisions & 5 & 6 & 7 & 8 & 9 & 10 & 11 & 12 \\
  \hline
  Number of Mazes & 2 & 9 & 5 & 16 & 13 & 10 & 4 & 1 \\ 
  A* length &  403 & 350 & 358 & 353 & 330 & 318 & 323 & 291 \\
  \hline\hline
  \end{tabular}
  \label{tab:maze_stats}
\end{table}

	Through the above process, more than 800 mazes were generated and tested but only 60 mazes were found to satisfy the two criteria. The 60 mazes used in this paper are available online\footnote{https://gitlab.com/2factor/QDMazes} and can be used as an open testbed for quality diversity algorithms. Details of the 60 mazes' properties are described in Table \ref{tab:maze_stats}, including the actual distance between the start and goal position based on A* pathfinding. It is immediately obvious that most mazes chosen have between 8 and 10 subdivisions, and that the A* distance decreases as the subdivisions increase. Indeed, there is a significant ($p < 0.01$) negative correlation between the number of subdivisions and the A* distance ($-0.39$). This is likely due to the fact that with more subdivisions the likelihood that a random maze would not be solvable even by NS-LC increases; for instance due to narrower corridors and more complex structures. Through an informal assessment of the many mazes generated, mazes with fewer subdivisions tend to fail the first criterion while mazes with more subdivisions tend to fail the second criterion.
	
	While the criterion of at least one success in 50 evolutionary runs for NS-LC is satisfied in all 60 mazes in our test set, it is worthwhile to investigate how many successes are actually scored by NS-LC per maze. The number of evolutionary runs (out of 50) in which NS-LC finds a solution is indicative of the hardness of the problem, and will be used in Section \ref{sec:experiments} as an important performance metric for comparing QD algorithms. Fig.~\ref{fig:maze_example} shows the 10 easiest mazes, where NS-LC found the most solutions in 50 runs. Among those 10 mazes, the average number of successes was $22.3$ (95\% CI $= 10.03$) while in all 60 mazes this was $5.93$ (95\% CI $= 2.45$). More broadly, the number of successes per maze ranged from 50 out of 50 (in 1 maze) to 1 out of 50 (in 20 mazes). There seems to be a significant ($p < 0.05$) correlation between NS-LC successes and A* path length (0.32). There is a weak negative correlation between NS-LC successes and number of subdivisions ($-0.11$) which is however not significant ($p > 0.05$). 
	
	\section{Experiments}\label{sec:experiments}
	
	As discussed in Section \ref{sec:maze_domain}, the maze navigation problem is used to test four quality diversity algorithms: NS-LC, SS-LC, NSS-LC and NS-SS-LC. Sixty generated mazes are used to test each algorithm in 50 evolutionary runs per method per maze with the same parameters: a population size of 250 individuals, a maximum of 600 generations, and a simulation time of 300 time steps.
	The core performance measure we consider is the aggregated number of evaluations per successful run across all mazes per method. A successful run discovers one robot that can reach the goal position from the start position within the 600 generations alloted; when a solution is found, evolution immediately ends. It is important to mention that for those methods where multi-objective optimization is applied (e.g., NS-LC), the final front of solutions is not considered as a performance indicator. Significance reported for all experiments in this paper is at a 95\% confidence. For multiple pairwise comparisons the Tukey's range test is used to establish significance. 
	
	In order to find appropriate values for the many parameters in each evolutionary method, a sensitivity analysis is performed on a subset of mazes and reported in Section \ref{sec:experiments_sensitivity}. The best parameters are selected for each algorithm which in turn are compared on their best setup in Section \ref{sec:experiments_results}. Note that evolutionary runs reported in Section \ref{sec:experiments_sensitivity} are independent from those reported in Section \ref{sec:experiments_results}.
	
	\subsection{Sensitivity Analysis}\label{sec:experiments_sensitivity}
	
	Due to a plethora of parameters that may impact the performance of QD algorithms examined in this paper, we rely largely on successful parameter values reported in the literature for the baseline algorithms. However, we perform a sensitivity analysis along two parameters for which we could not find suggested values: the locality of local competition ($n_{LC}$) in all QD algorithms and the weight of novelty versus surprise ($\lambda$) in NSS-LC. Since most mazes in the test set are particularly difficult to solve, the impact of a parameter tweak is not expected to have a strong impact on the algorithm's performance\footnote{For example, in a maze where NS-LC has only one success in 50 runs, the impact of any parameter is expected to be a product of chance.}. Therefore, the 10 easiest mazes (based on ``vanilla'' NS-LC) shown in Fig.~\ref{fig:maze_example} are used for the sensitivity analysis in this paper. The core performance metric for sensitivity analysis is the average number of evaluations needed by each algorithm to discover a controller able to solve the maze. In a run where no solution was found within the allocated budget, the maximum allocated evaluations ($150{\cdot}10^3$) is used. This metric is averaged across 50 evolutionary runs.
	
	Beyond testing the sensitivity of $n_{LC}$ and $\lambda$, which we discuss in Section \ref{sec:experiments_sensitivity_parameters}, we assess how the various sub-components of the introduced algorithms (novelty, surprise and the linear combination of novelty and surprise) affect algorithmic performance with an additional set of baseline algorithms described in Section \ref{sec:experiments_sensitivity_components}. Section \ref{sec:experiments_sensitivity_parameters} reports results based on 50 evolutionary runs per maze; Section \ref{sec:experiments_sensitivity_components} also reports results based on 50 evolutionary runs per maze, which are independent from those of Section \ref{sec:experiments_sensitivity_parameters}. Finally, experiments in Section \ref{sec:experiments_results} perform evolutionary runs independently from either Section \ref{sec:experiments_sensitivity_parameters} or \ref{sec:experiments_sensitivity_components}. 
	
\subsubsection{Sensitivity to Parameters}\label{sec:experiments_sensitivity_parameters}
	
\begin{figure}[!tb]
    \centering
    \subfloat[Local competition $n_{LC}= 5$]{
    \includegraphics[width=0.48\linewidth]{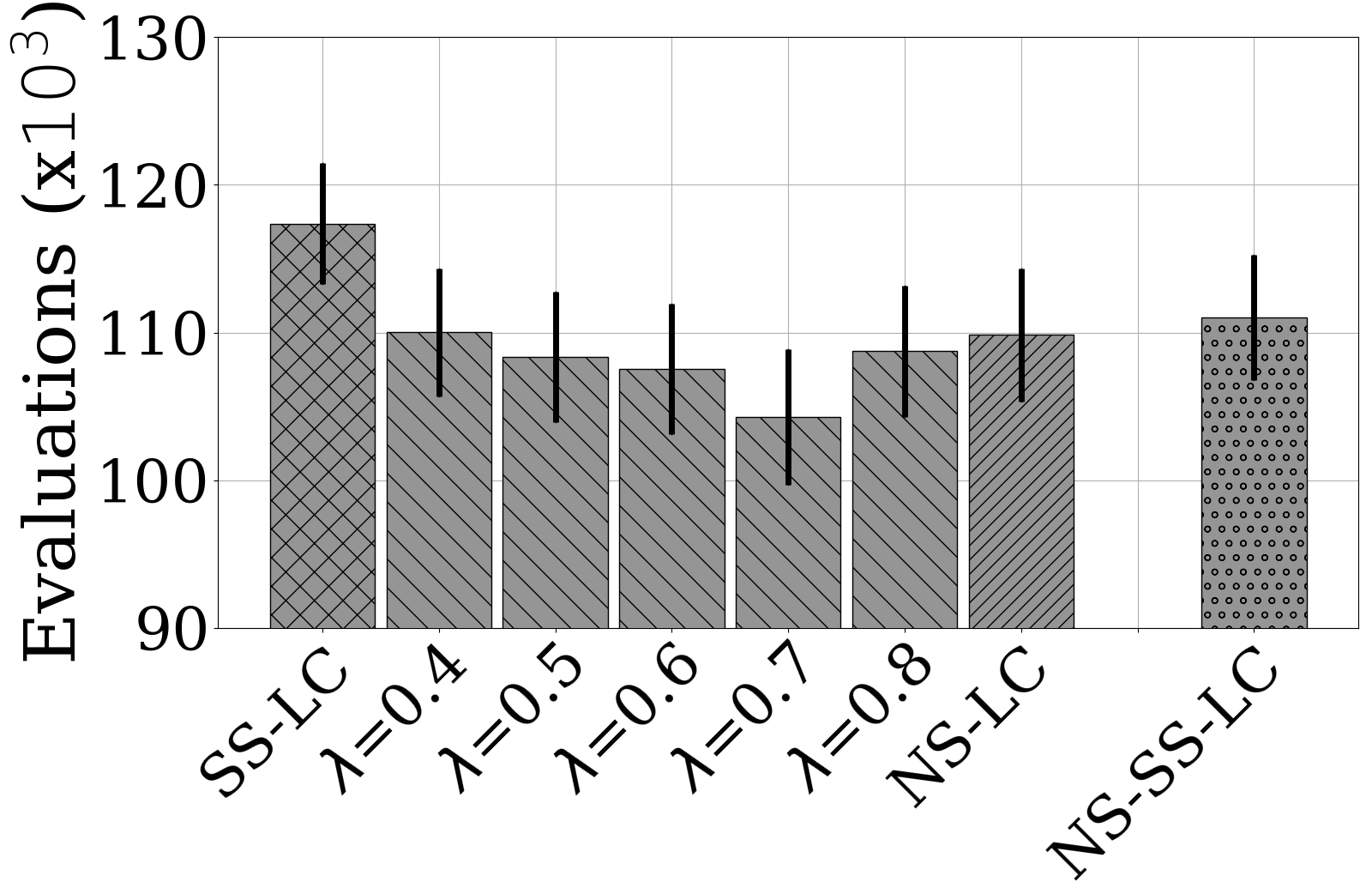}
    \label{fig:sensitivity_lc_5}}
    \subfloat[Local competition $n_{LC}= 10$]{
    \includegraphics[width=0.48\linewidth]{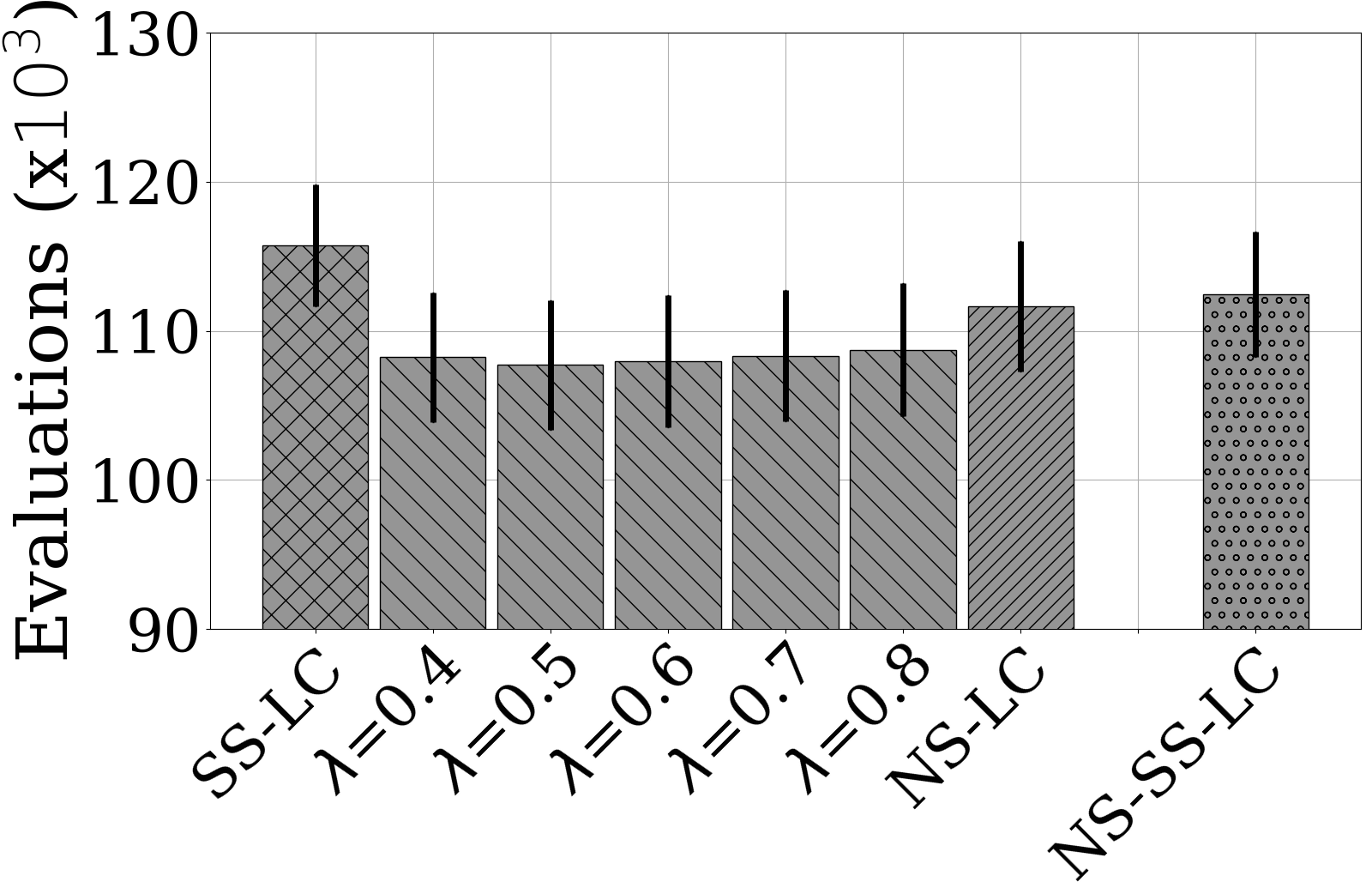}
    \label{fig:sensitivity_lc_10}} 
    \\
    \subfloat[Local competition $n_{LC}= 15$]{
    \includegraphics[width=0.48\linewidth]{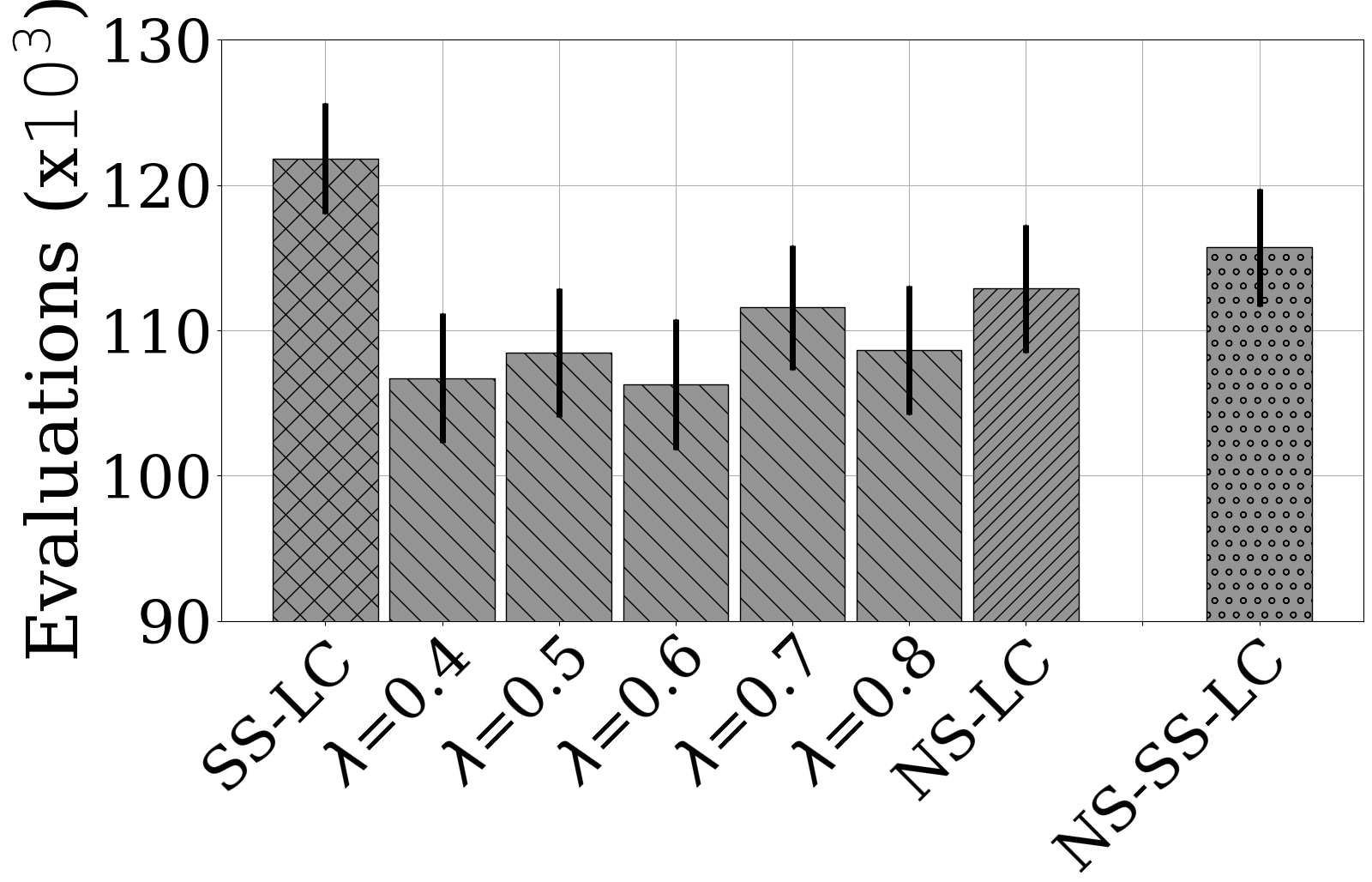}
    \label{fig:sensitivity_lc_15}}
    \subfloat[Local competition $n_{LC}= 20$]{
    \includegraphics[width=0.48\linewidth]{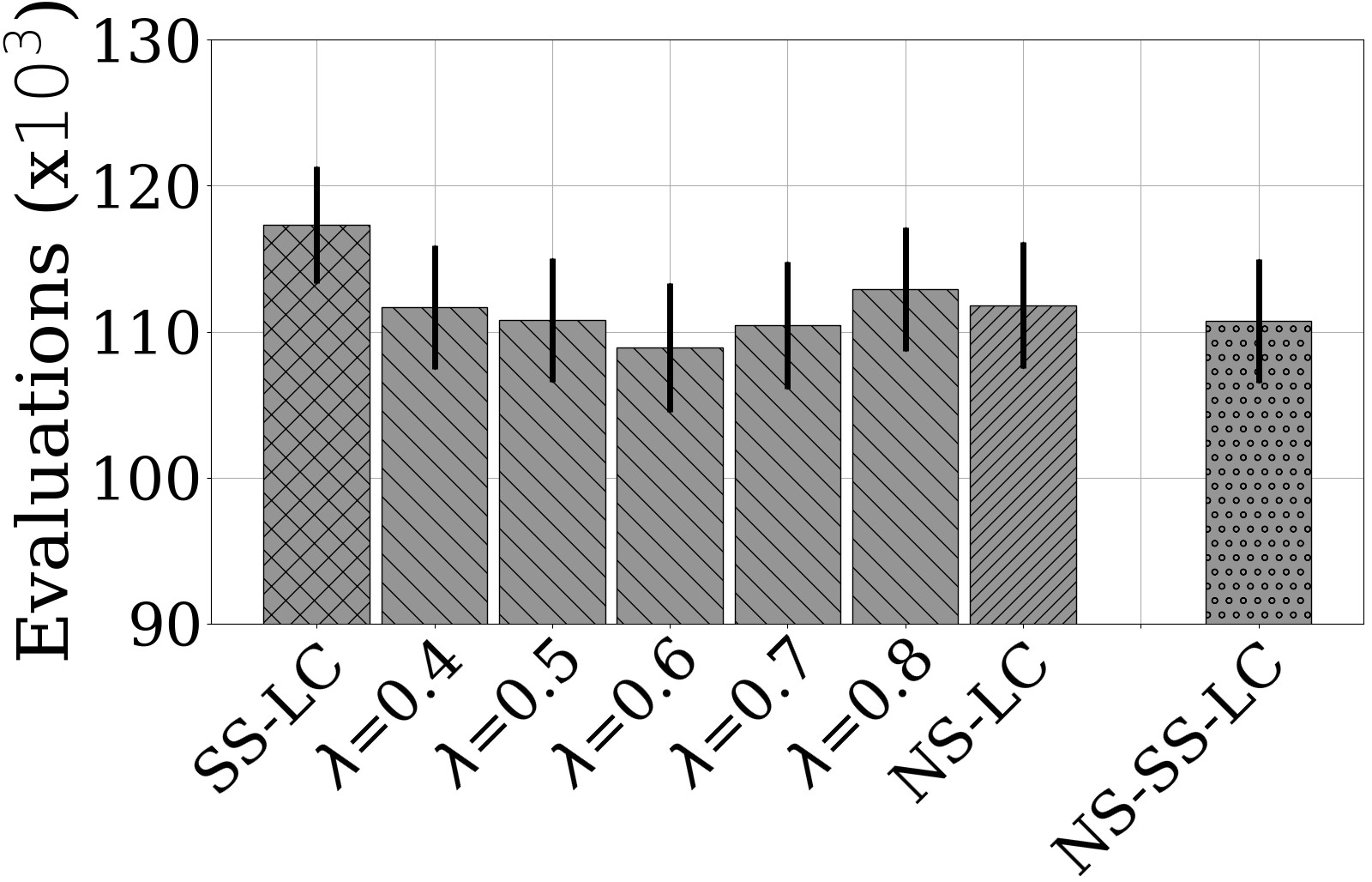}
    \label{fig:sensitivity_lc_20}}

\caption{\textbf{Sensitivity analysis:}
Average number of evaluations in the 10 easiest mazes across four local competition sizes (5, 10, 15, 20) and five values of $\lambda$ (0.4, 0.5, 0.6, 0.7, 0.8). Error bars denote 95\% confidence intervals.}
\label{fig:sensitivity}
\end{figure}

The parameter setups range from a $n_{LC}$ value between 5 and 20 (increments of 5) and $\lambda$ for NSS-LC between 0.4 and 0.8 (increments of 0.1). The results of this analysis in terms of average number of evaluations across all runs in all 10 easiest mazes are shown in Fig.~\ref{fig:sensitivity}.

	The results of the sensitivity analysis show that both parameters have an impact on the performance of both SS-LC and NSS-LC (where applicable). Notably, however, $n_{LC}$ seems to have very limited impact on the performance of NS-LC, with marginally better performance obtained with $n_{LC}=5$. The same parameter impacts SS-LC in a much more pronounced manner, with $n_{LC}=5$ and $n_{LC}=15$ giving a far worse performance for SS-LC than other values tested. The best performance for SS-LC is for $n_{LC}=10$, which however requires more evaluations than NS-LC. NS-SS-LC tends to fluctuate in the same way as NS-LC, although it seems to fall behind for $n_{LC}=15$. Finally, NSS-LC hinges on both $\lambda$ and $n_{LC}$ and it is clear from Fig.~\ref{fig:sensitivity} that good performance can be achieved for several combinations of these two parameters. However, the least evaluations are achieved with a high $\lambda$ value (see Table \ref{tab:final_parameters}), in which case the novelty score contributes more than the surprise score. 
	Based on this analysis, the parameters for the remaining experiments in this paper (Sections \ref{sec:experiments_sensitivity_components} and \ref{sec:experiments_results}) are shown in Table \ref{tab:final_parameters}.

\begin{table}
\centering
\caption{Chosen parameters for QD algorithms based on a sensitivity analysis, and the average number of evaluations (with 95\% confidence intervals) obtained on the mazes of Fig.~\ref{fig:maze_example}.}
\label{tab:final_parameters}
\begin{tabular}{l||p{3.9cm}l}
\hline\hline
Algorithm & Parameters & Evaluations (${\cdot}10^3$) \\ \hline\hline
NS-LC & $n_{LC}=5$, $n_{NS}=15$ & $109.8 \pm 4.4$ \\ \hline
\multirow{2}{*}{SS-LC} & $n_{LC}=10$, $n_{SS}=2$,  & \multirow{2}{*}{$115.7 \pm 4.0$}\\ 
& $h=2$, $k_{SS}=200$  & \\ \hline
\multirow{2}{*}{NSS-LC} & $\lambda=0.7$, $n_{LC}=5$, $n_{NS}=15$, & \multirow{2}{*}{$104.2 \pm 4.5$} \\ 
&  $n_{SS}=2$, $h=2$, $k_{SS}=200$ & \\ \hline
\multirow{2}{*}{NS-SS-LC} & $n_{LC}=5$, $n_{NS}=15$, $n_{SS}=2$, & \multirow{2}{*}{$111.0 \pm 4.2$} \\ 
& $h=2$, $k_{SS}=200$ & \\ \hline\hline
\end{tabular}
\end{table}
		
	\subsubsection{Sensitivity to Algorithmic Components}\label{sec:experiments_sensitivity_components}
	
	Apart from the many different parameters in the proposed QD algorithms, there are several design decisions in the implementation of each. Based on the core components of algorithms considered (novelty, surprise, local competition), we will compare the four QD algorithms against novelty search (described in Section \ref{sec:maze_domain_algorithms}) and four more baselines: 
	\begin{itemize}
		\item Surprise Search (SS), described in Section \ref{sec:algorithms_divergence_surprise}. Surprise search uses the same predictive model as SS-LC, as described in Section \ref{sec:maze_domain_algorithms} and Fig.~\ref{fig:surprise_search}. The parameters of SS ($n_{SS}$, $h$, $k_{SS}$) are the same as SS-LC in Table \ref{tab:final_parameters}.
		\item Novelty-Surprise Search (NSS), which is a single objective implementation that linearly combines novelty and surprise (see Eq. \ref{eq:novelty_surprise}) as described in \cite{gravina2017noveltysurprise}. Performing the same analysis across $\lambda$ values as in Section \ref{sec:experiments_sensitivity_parameters}, we use $\lambda=0.4$ for NSS while $n_{SS}$, $h$, and $k_{SS}$ are the same as NSS-LC in Table \ref{tab:final_parameters}.
		\item Novelty Search--Surprise Search (NS-SS) uses a steady-state NSGA-II multi-objective algorithm \cite{li2017efficient} to search for non-dominated solutions on the dimensions of novelty (Eq.~\ref{eq:novelty}) and surprise (Eq.~\ref{eq:surprise}). All parameters are the same as NS-SS-LC in Table \ref{tab:final_parameters}.
		\item Surprise Search Archive with Local Competition (SSA-LC), which is a variant of SS-LC where the divergence objective of NSGA-II uses the surprise score but also maintains a novelty archive identical to how it is maintained in NS-LC \cite{lehman2011creatures} (using the same fluctuating threshold and assessing individuals for insertion to the novelty archive based on the novelty score of Eq.~\ref{eq:novelty}). To calculate the surprise score, SSA-LC considers the nearest neighbors both in the prediction space and in the novelty archive. To calculate the local competition score, SSA-LC considers the nearest neighbors in the current population and in the novelty archive (similar to NS-LC). Results for the best performing $n_{LC}$ parameter are reported ($n_{LC}=5$), while other parameters of surprise search ($n_{SS}$, $h$, $k_{SS}$) are the same as SS-LC in Table \ref{tab:final_parameters}.
	\end{itemize}
	
	The first three baseline algorithms test only the diversity dimensions of some of the proposed QD algorithms (NSS is the diversity dimension used in NSS-LC, while NS-SS is the two-objective diversity component of NS-SS-LC). SSA-LC tests another way of combining deviation from expected (through the prediction space) and seen (through the novelty archive) behaviors.

\begin{figure}[tb]
\centering
\includegraphics[height=0.55\linewidth]{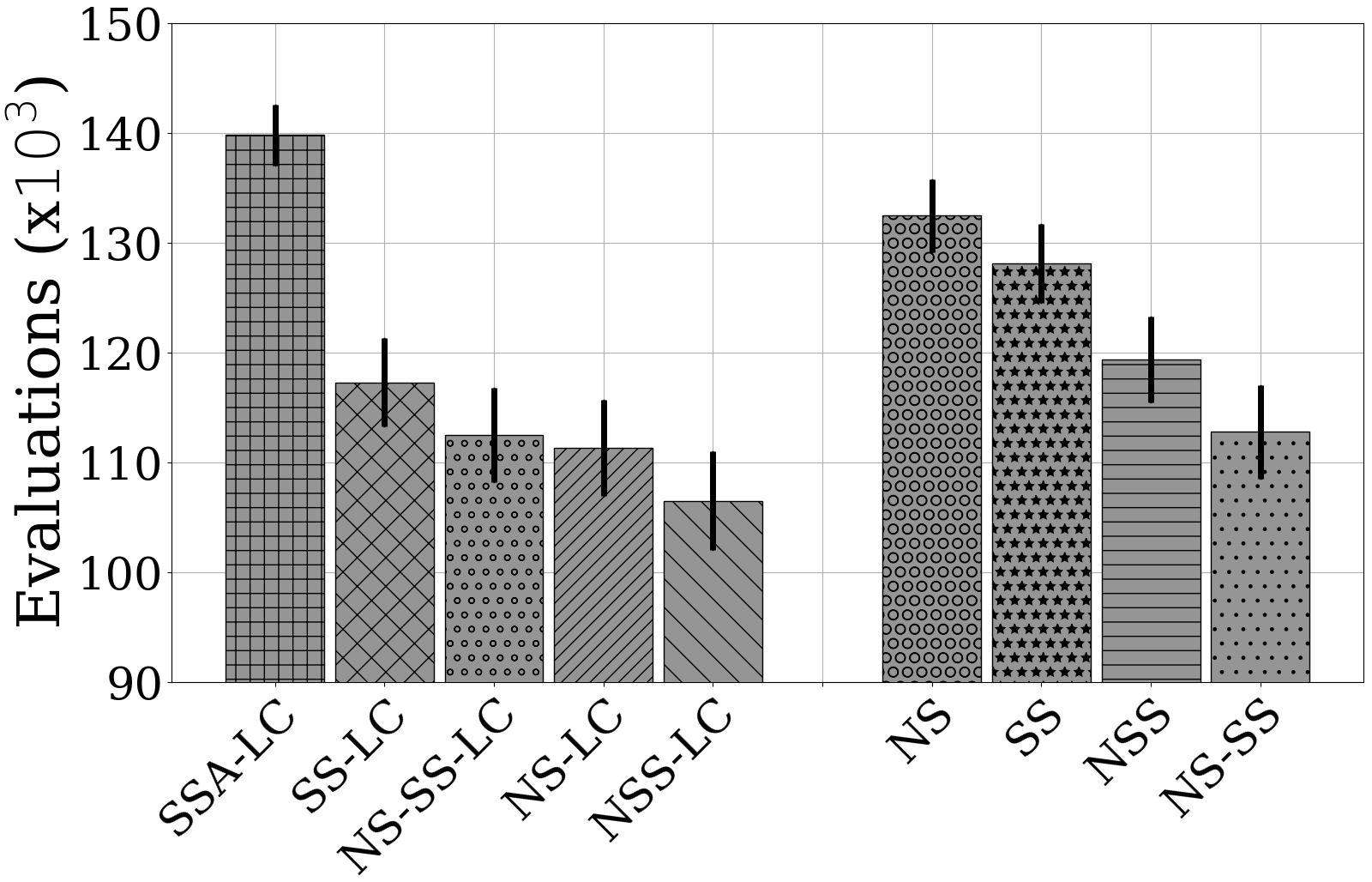}
\caption{\textbf{Sensitivity to algorithmic components:} Average number of evaluations in the 10 easiest mazes across five QD approaches (NS-LC, NSS-LC, SS-LC, SSA-LC, NS-SS-LC) and four divergent search algorithms (NS, SS, NSS, NS-SS). Error bars denote 95\% confidence intervals.}
\label{fig:sensitivity_components}
\end{figure}
	
	Fig.~\ref{fig:sensitivity_components} shows the performance of all algorithms discussed in this paper, separating QD approaches (on the left of the figure) from pure divergent search approaches (on the right). It is clear that using a local competition as a second objective significantly improves performance over a divergence-only variant (e.g., SS-LC versus SS, NSS-LC versus NSS). Notably, the NS-SS multi-objective divergent search approach performs surprisingly well, being the most efficient divergent search approach and requiring significantly fewer evaluations than both SS and NS. NS-SS actually performs at a similar level as some QD approaches (such as SS-LC).
	
	Fig.~\ref{fig:sensitivity_components} also highlights that while novelty search alone underperforms compared to surprise search, NSS and SS-NS, when combined into a QD algorithm the performance rankings are much different. NSS-LC and NS-LC perform much better compared to SS-LC (significantly for NSS-LC), although the fact that a version of NS-LC was used to find appropriate mazes that it can solve may have been a reason for the good performance of NS-LC. Moreover, NS-SS-LC does not perform better than NSS-LC and NS-LC in the same way as NS-SS does over NSS and NS. While the performance of NS-SS-LC is further discussed in Sections \ref{sec:experiments_results} and \ref{sec:discussion}, a likely reason is that the simultaneous optimization of three objectives makes the problem more difficult for a multi-objective approach to solve \cite{purshouse2007conflicting}.
	
	Finally, it is obvious that the introduction of a novelty archive to surprise search leads to a much worse performance. The working hypothesis for this behavior is the divergence dimension of this multi-objective approach. Specifically, ``vanilla'' surprise search attempts to deviate from predicted points in the search space, which can force it to back-track to previously seen areas of the search space. Introducing a novelty archive, where such previously visited areas of the search space are kept,
	actively deters back-tracking and impairs the surprise search component of SSA-LC.
	
	\subsection{Comparison of QD Algorithms}\label{sec:experiments_results}
	
	Having ensured an optimal set of parameters for all QD algorithms tested, each of the four quality diversity methods are applied to each of the 60 generated mazes. Comparisons among QD methods---and against novelty search as an indicative divergent search algorithm---aim to answer two core questions: whether an algorithm is more likely to find a solution in an evolutionary run than others (addressed in Section \ref{sec:experiments_results_successes}), and whether an algorithm finds such a solution in fewer evaluations (in Section \ref{sec:experiments_results_robustness}). While not a measure of performance, the evolved controllers are also compared in terms of the complexity of the ANN evolved (in Section \ref{sec:experiments_results_genome}), as an indication of genotypes favored by each approach.
	
	\subsubsection{Successes}\label{sec:experiments_results_successes}
	
	Table \ref{tab:tournament_successes} shows how each method compares in terms of number of successes in each of the 60 generated mazes. In this table, the number of mazes where one approach outperforms the other (in terms of successful runs out of 50) is shown on a per column and per row basis. As expected, novelty search as a pure divergent approach has the lowest rank compared to the other algorithms with only 2\% instances outperforming the four QD algorithms on average and 79\% instances being outperformed. On the other hand, NSS-LC outperforms all approaches more often (59\%) than it is outperformed (20\%). Notably, SS-LC does not perform equally well, as it is often outperformed by NS-LC (52\%); however, NSS-LC outperforms NS-LC (48\%) more often than it is outperformed by NS-LC (30\%). It is interesting that NS-SS-LC outperforms NS-LC (42\%) marginally more often than the opposite (40\%), but when compared to all the approaches it has fewer instances of superior behavior (49\%) than NS-LC.
	
	It is important to note that NSS-LC is also superior to NS-LC in the 10 easiest mazes, where both approaches are more likely to have a successful run than in the hardest mazes of the test set; in harder mazes, a successful run is largely a matter of chance. In the 10 easiest mazes of Fig.~\ref{fig:maze_example}, NSS-LC outperforms NS-LC in 6 of 10 mazes and is outperformed by NS-LC in 4 mazes. Note that results in Table \ref{table:rankings} are from 50 runs independent of the runs performed for the sensitivity analysis of Section \ref{sec:experiments_sensitivity}; however, admittedly the parameters of both NS-LC and NSS-LC were optimized explicitly for these 10 mazes of Fig.~\ref{fig:maze_example}.
\begin{table}[!tb]
\centering
\caption{Algorithms tournament: Percentage of 60 generated mazes for which the algorithm in a row has a strictly greater (${\geq}1$)  number of successes compared to the algorithm in a column. Last row and last column are respectively the average of each column and the average of each row.}\label{table:rankings}
\begin{tabular}{l@{ }||@{ }c@{ }|@{ }c@{ }|@{ }c@{ }|@{ }c@{ }|@{ }c@{ }||@{ }c@{ }}
\hline\hline    
& NS & NS-LC & NSS-LC & SS-LC & NS-SS-LC &  Average    \\
\hline \hline     
NS & -- & 1.7\% & 0.0\% & 5.0\% & 1.7\% & 2.1\% \\ \hline
NS-LC & 83.3\% & -- & 30.0\% & 51.7\% & 40.0\% & 51.2\% \\ \hline
NSS-LC & 81.7\% & 48.3\% & -- & 60.0\% & 45.0\% & 58.7\% \\ \hline
SS-LC & 70.0\% & 28.3\% & 25.0\% & -- & 30.0\% & 38.3\% \\ \hline
NS-SS-LC & 80.0\% & 41.7\% & 26.7\% & 46.7\% & -- & 48.7\% \\
\hline \hline					
Average & 78.7\% & 30.0\% & 20.4\% & 40.8\% & 29.2\% & -- \\
\hline\hline
\end{tabular}
\label{tab:tournament_successes}
\end{table}
	
	\subsubsection{Robustness}\label{sec:experiments_results_robustness}
\begin{figure}[!tb]
		\centering
		\includegraphics[height=0.64\linewidth]{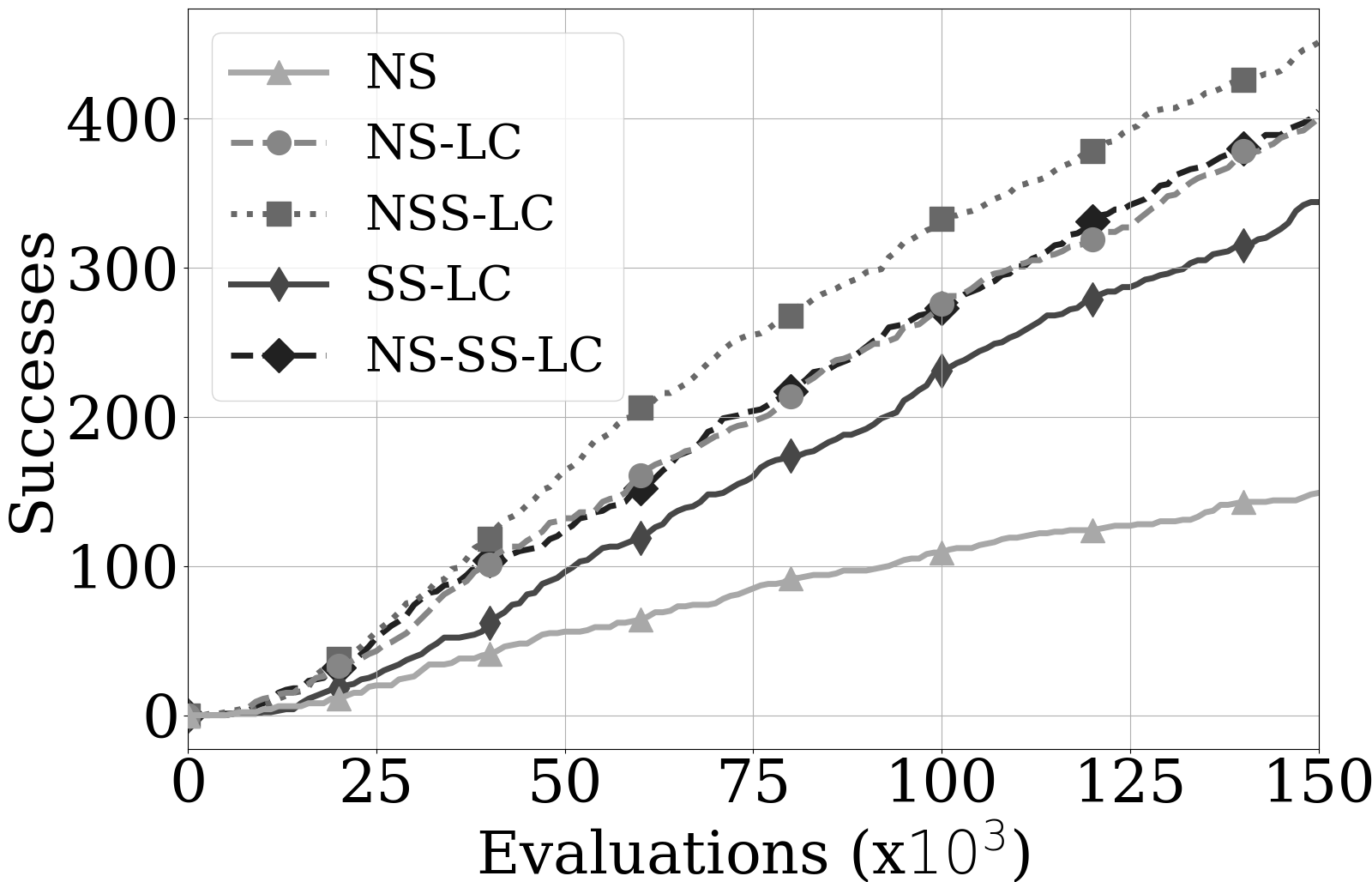}
		\caption{\textbf{Robustness:} number of successes over evaluations by aggregating all the runs across the 60 generated mazes for each approach.}
		\label{fig:robustness}
\end{figure}
    
\begin{figure}[!tb]
\centering
\includegraphics[height=0.6\linewidth]{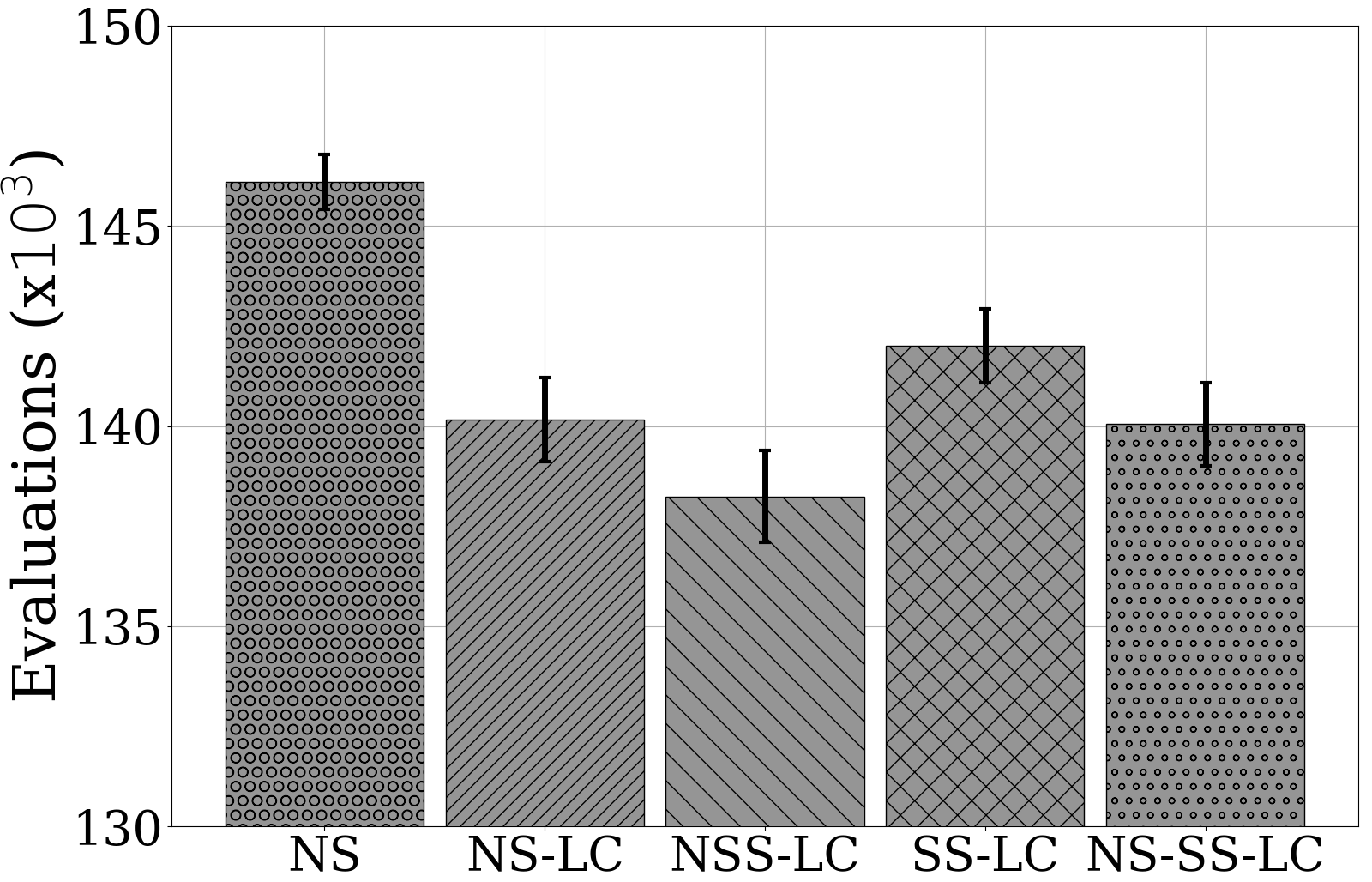}
\caption{\textbf{Average number of evaluations} from all the runs across the 60 generated mazes of each algorithm (i.e., 3000 runs per algorithm). Error bars denote 95\% confidence intervals.}
\label{fig:evaluations}
\end{figure}

	While the number of evolutionary runs resulting in a success is a good indication of algorithms' performance, it is worthwhile to investigate how quickly those solutions are found. We use robustness as a secondary performance metric: robustness is defined as the number of overall successes obtained by each algorithm after a number of evaluations. This indicates how consistent the results are, should evolution stop earlier than the current generation limit.
	
	Fig. \ref{fig:robustness} and \ref{fig:evaluations} report the robustness and the average number of evaluations respectively per method. Results are aggregated among all the runs of the 60 generated mazes per method, i.e., a total of 3000 runs. Aggregating successes across mazes allows for a more realistic view of algorithmic performance: averaging successes across the 60 mazes leads to very large deviations as some mazes are much harder than others (based on NS-LC performance described in Section \ref{sec:maze_domain_generation}). Unsurprisingly, novelty search is outperformed by every other algorithm from $20{\cdot}10^3$ evaluations onwards, achieving a total of 149 successful runs in 60 mazes. NSS-LC outperforms NS-LC and NS-SS-LC after $20{\cdot}10^3$ evaluations and consistently reaches more successes than any other method from that point onward. When the maximum evaluations are reached, NSS-LC obtains 451 successes versus the 400 successes of NS-LC, 405 successes of NS-SS-LC and the 344 successes of SS-LC.
	The superiority of NSS-LC over NS-LC is also evident in Fig.~\ref{fig:evaluations}, as on average NSS-LC requires significantly fewer evaluations than every other algorithm. This finding is not surprising per se, since NSS-LC finds more solutions while other approaches often spend their entire budget (evaluations) searching but not finding a solution. However, Fig.~\ref{fig:evaluations} establishes that this difference in successes results in a statistically significant acceleration of the algorithm. NS-SS-LC seems to perform very similarly to NS-LC, both in terms of number of successes over the progress of evolution, and in terms of average evaluations of Fig.~\ref{fig:evaluations}. This observation is corroborated by tests in Sections \ref{sec:experiments_sensitivity_parameters} and \ref{sec:experiments_sensitivity_components}. Finally, results of SS-LC are underwhelming also in terms of robustness, as it quickly falls behind both NSS-LC and NS-LC, and on average it needs significantly more evaluations than other QD alternatives.

	\subsubsection{Genomic Complexity}\label{sec:experiments_results_genome}
	
\begin{figure}[tb]
\centering
\subfloat{\includegraphics[height=0.31\linewidth]{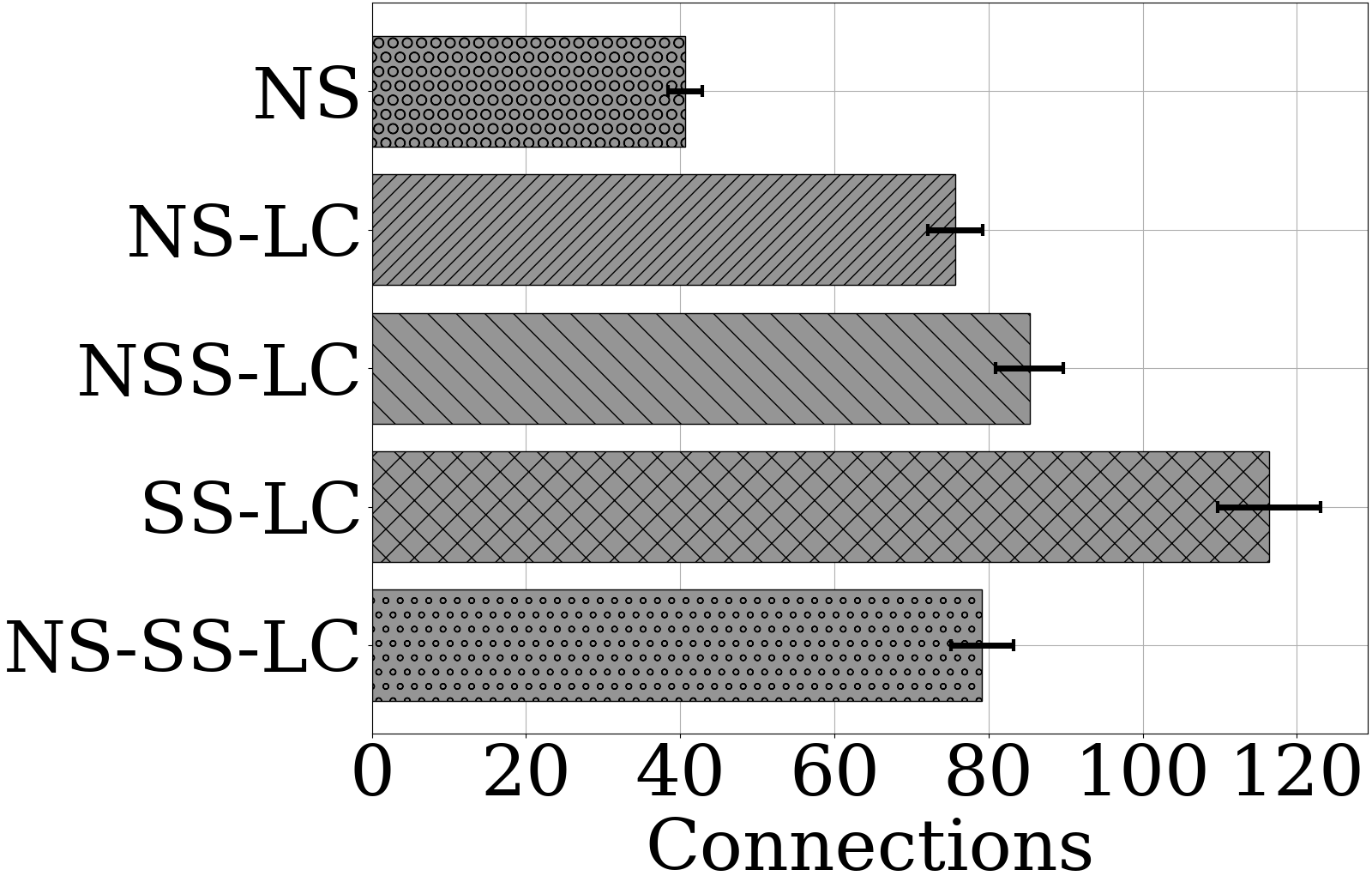}} 
\hfill
\subfloat{\includegraphics[height=0.31\linewidth]{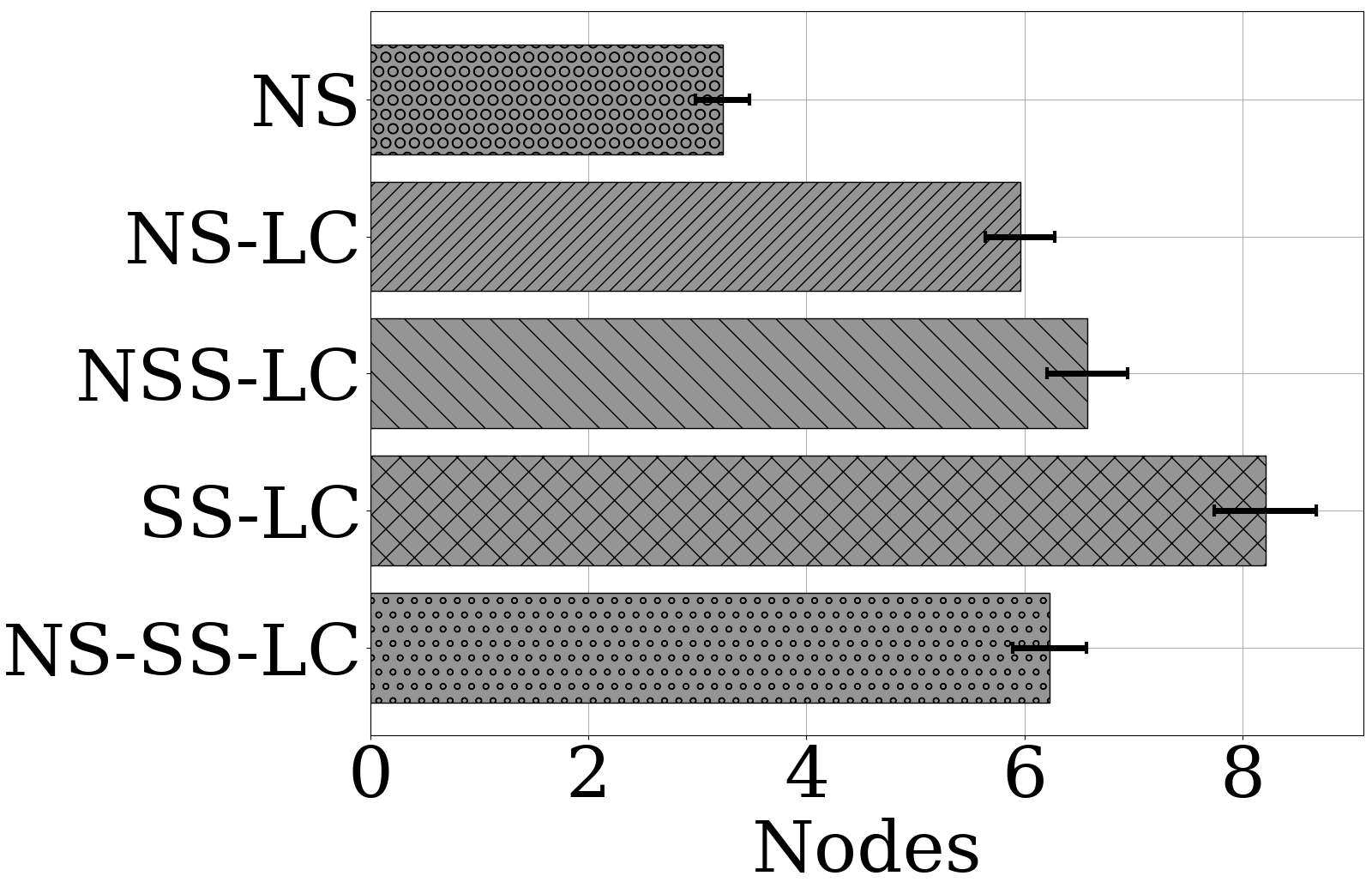}}
\caption{\textbf{Complexity:} number of connections (left figure) and hidden nodes (right figure) on average for evolved ANNs which solve the mazes per approach. Error bars denote 95\% confidence intervals.}
\label{fig:complexity}
\end{figure}

	As an indication of the type of solutions favored by each evolutionary method, we compare the complexity of the evolved ANNs which were successful in finding the goal in each method (i.e., 400 ANNs for NS-LC, 149 ANNs for novelty search etc.). Genotypic complexity refers here to the average number of hidden nodes and connections of a successful ANN. In \cite{gravina2016surprisebeyond, gravina2017noveltysurprise, gravina2017surprise}, the genotypic complexity across three divergent search methods (novelty, surprise and novelty-surprise search) showed that each method favored different structures, with surprise search favoring far larger networks than novelty search.
	Figure \ref{fig:complexity} shows the average number of hidden nodes and connections per approach. As in previous findings \cite{gravina2016surprisebeyond}, novelty search favors very small networks (few connections, few hidden nodes), while NS-LC has denser and larger networks than novelty search but less than both NSS-LC and SS-LC (significantly so for SS-LC). Finally, it is interesting to note that NSS-LC has significantly smaller and sparser networks than SS-LC, striking a happy (i.e., most effective) medium between NS-LC and SS-LC. Structural metrics of ANNs evolved by NS-SS-LC have no significant differences with either NSS-LC or NS-LC, but seem to fall between the two; the similarity with networks of NS-LC is further evidence that NS-SS-LC performs a very similar search process as NS-LC, which explains their similar performance in Fig.~\ref{fig:robustness}.

	\section{Discussion}\label{sec:discussion}
	
	In the experiments of Section \ref{sec:experiments}, we compared three new quality diversity algorithms with alternate dimensions of diversity combined with local competition in terms of robustness and efficiency in discovering solutions in diverse maze setups. The key findings are that NSS-LC outperforms the other QD approaches considered in this work, namely NS-LC, NS-SS-LC and SS-LC; NS-SS-LC has a very similar performance as NS-LC, while SS-LC underperforms by comparison. The findings on NSS-LC validate our hypothesis that surprise can be a beneficial form of search for quality diversity. Coupled with our findings from the sensitivity analysis in Section \ref{sec:experiments_sensitivity} when components of the QD algorithms were added or removed, we speculate and discuss two likely reasons why adding surprise to novelty helps to achieve better overall performance.
	
	A core hypothesis for the differences in algorithmic performance is that surprise search allows the algorithm to back-track, re-visiting areas of the search space that have been explored in previous generations. Since surprise search operates on diverging from predictions, it may favor past behaviors if they are sufficiently different from predicted future trends. We have already provided such an example from nature in Section \ref{sec:introduction}, where ``terror birds'' were surprising as they went against behavioral trends (birds were becoming smaller and better fliers) while not being particularly novel historically. By comparison, novelty search actively discourages re-visiting areas of the search space due to the novelty archive. This has been a hypothesis for the efficiency of surprise search compared to novelty search in purely deceptive tasks \cite{gravina2016surprisebeyond}, although admittedly it is difficult to prove quantitatively. However, an indication of the back-tracking nature of surprise is gleaned by the more complex networks evolved to solve the mazes in Table \ref{fig:complexity}; the increased size compared to novelty search, which is consistent in other studies \cite{gravina2017noveltysurprise,gravina2016surprisebeyond}, is likely because same areas of the behavioral space are re-visited in later stages of evolution with larger networks. A clearer indication is that when surprise search with local competition also considers an archive of past novel individuals (SSA-LC), its performance drops significantly. We expect that the influence of an archive which actively deters back-tracking goes against the main drive of surprise search. It is evident that the back-tracking of surprise search does not seem to perform well in QD algorithms, as combining it with local competition seems to lead to local optima. This is perhaps why SS-LC underperforms in these tests, although it should be noted that it still scores more successes than novelty search alone in 70\% of mazes. On the other hand, allowing novelty search to back-track through a surprise score component is likely why NSS-LC performs better. There is a fine balance between too much back-tracking (e.g., in SS-LC) and too little (e.g., in NS-LC), but the fact that NSS-LC performs better at higher $\lambda$ values (e.g., $\lambda=0.6$ or $\lambda=0.7$ in Fig.~\ref{fig:sensitivity}) should be an indication that the novelty score primarily drives the search and back-tracking is favored in specific circumstances.
	
	The second likely cause for a high-performing NSS-LC is based on recent research on novelty in highly successful patents, where the authors focused on the ``sweet spot'' of novelty \cite{he2017novelty}. That study suggests that a sweet spot of novelty might exist on the border of what we can call a ``conventional solution'' and a completely novel solution. A similar hypothesis suggests that optimal solutions exist on the border between feasible and infeasible spaces in constrained optimization \cite{schoenauer1996edge}. In the maze navigation domain, for instance, behaviors that are ``too novel'' might dominate solutions that are less novel but more promising. Surprise may help to search in a more fine-grained fashion the space between a completely new solution and solutions already in the population or in the archive, in order to find the desired sweet spot of novelty. This is not only possible through back-tracking, but especially in NSS it is due to the balance of the two scores (novelty versus surprise). When the population in one generation becomes too novel compared to that of the previous generation, predictions will be even more distant points; in this case, the surprise score will have a larger impact as the distance from predicted points will be larger than that of past or current points. In general this balancing factor softens the greedy search for all-new solutions and, coupled with the secondary objective of local competition, results in a more efficient search process.

The difference between NS-SS-LC and NSS-LC is perhaps the most surprising, as we would expect the decoupling of novelty and surprise scores to perform better compared to an aggregated approach. The very good performance of NS-SS as a two-objective divergent search algorithm (outperforming NSS) also indicated that a decoupling of the two measures of divergence would be beneficial and pointed to the orthogonal nature of surprise and novelty. However, as a QD algorithm the three-objective NS-SS-LC approach seemed to perform on par with NS-LC, to the degree of evolving similar sized networks. We can assume that the simultaneous search for three dimensions (compared to all other QD approaches which search along two dimensions) was the primary cause of its subdued performance. As dimensions increase, so does the number of non-dominated solutions \cite{purshouse2007conflicting}, which makes the search process slower. A further complication is that all three of the dimensions have dynamic, fluctuating scores: novelty search is sensitive to both the population and the novelty archive and the same individual may receive a different novelty score from one generation to the next. The same applies for surprise score, which depends on the ever-changing recent trends and predictions which are recalculated in every generation, and the local competition which again depends on other individuals in the population and the novelty archive. How multi-objective algorithms handle such dynamic fitness dimensions has not been sufficiently examined, and is an interesting direction of inquiry. Due to the high performance of NS-SS, however, we expect that an algorithm that can in principle handle more objectives more efficiently will lead to improved performance for NS-SS-LC. We can envision that a reference point based algorithm, such as NSGA-III \cite{deb2014evolutionary}, might mitigate some of these problems and will be considered for future work.

We should note that all algorithms in this work use a steady-state version of NEAT, as in \cite{lehman2011abandoning}. In particular, the multi-objective algorithms (NS-LC, SS-LC, NSS-LC, NS-SS-LC) use an improved version of NSGA-II for steady-state implementations \cite{li2017efficient}. There are three main reasons behind this choice: first, for fair comparisons with the baselines used in this work (NS, SS, NSS) which use a steady state implementation for the maze navigation testbed \cite{lehman2011abandoning, gravina2016surprisebeyond, gravina2017noveltysurprise}; second, due to evidence in \cite{lehman2013diversity} that the generational counterpart of novelty search does not perform equally well in a maze navigation scenario, likely due to a ``less informative gradient'' for novelty search given by a generational reproduction mechanism \cite{lehman2013effective}; third, due to arguments in recent studies \cite{durillo2009effect, nebro2009effect, buzdalov2015various} that a steady-state multi-objective implementation can be beneficial in terms of convergence and diversity for particular problems. Based on these studies we assume that a steady state implementation is more suitable for the maze navigation testbed. Nevertheless, future work will test the degree to which this assumption holds by comparing current findings against a generational implementation of the proposed QD algorithms.

\subsection{Extensions and Other Applications}
	
This paper focused on enhancing one QD approach: a multi-objective blend of a divergence score and a measure of local superiority. Other approaches for QD such as MAP-Elites \cite{mouret2015illuminating} could also be considered, either as another QD algorithm used as a benchmark for NSS-LC, or to actually introduce surprise as a way of searching within the space of MAP-Elites. Surprise could be integrated into MAP-Elites, for instance, by storing two different maps: one used in the normal algorithm and the other used for prediction, with deviation from predictions used as a selection process for the algorithm.
	
An important direction for surprise-based QD algorithms is exploring different prediction models ($m$ in eq.~\ref{eq:prediction_model}) for the computation of surprise. While a simple linear regression model seems to be sufficient for delivering state of the art results in the maze navigation domain, other surrogate models \cite{jin2005comprehensive} such as \emph{kriging} modeling could be investigated for more complex tasks and domains \cite{gaier2017data}.
	
Searching for quality diversity in other domains is also a promising broad area for future work. Quality diversity is particularly well-suited, for example, in procedural content generation (PCG). PCG algorithms are used to create content for games, such as levels, buildings, weapons or characters \cite{shaker2016procedural}. However, such content needs to fulfill several desired properties, such as high quality, reliability, believability and diversity. Generating content that shows all these desired properties at the same time is difficult; most PCG algorithms focus currently only on one dimension. For example, if we focus on quality and diversity of the generated content, a trade-off needs to be made. If the search space is too restricted, quality can be guaranteed, but the solutions will be more similar. On the other hand, pushing for diversity can lead to poorly suboptimal solutions or---if the search space is too broad---unusable content. While this problem has already been tackled successfully in different ways \cite{preuss2014gooddiverse, liapis2015ecj, gravina2016constrained}, an innovative take would be to introduce the proposed QD algorithms for PCG, thereby testing their capabilities both in terms of diversity and quality of the artefacts generated. In fact, while in constrained optimization PCG approaches \cite{liapis2015ecj, gravina2016constrained} the search space is subdivided into feasible and infeasible space, a QD approach might discover more diverse solutions as the search space is divided into multiple niches. Finally, the surprise-based QD algorithms should be tested against a broader set of state of the art evolutionary algorithms, such as CMA-ES \cite{hansen2006cma} or multimodal optimization algorithms \cite{preuss2015multimodal}.
	
\section{Conclusion}\label{sec:conclusion}
	
This paper explored the impact of different diversity strategies on quality diversity evolutionary approaches and tested the hypothesis that surprise search may augment the exploration capacity of QD algorithms. Building on the concept of novelty search as one dimension coupled with local competition as a second dimension, alternatives to novelty search which used surprise or a combination of novelty and surprise were devised. These new QD algorithms were tested extensively on the maze navigation domain; we used a broad set of 60 mazes with varying degrees of deceptive fitness landscapes and each algorithm was evaluated across a total of 3000 evolutionary runs. Experiments concluded that an aggregated novelty-surprise search with local competition outperforms other QD algorithms, both in terms of number of runs which find a solution to the problem and the number of evaluations in which such solutions are found. These findings open new directions in exploring different notions of divergence, and establish surprise as an alternative notion to novelty not only for divergent search but also for quality diversity algorithms.

\section*{Acknowledgments}
	
This work has been supported in part by the FP7 Marie Curie CIG project AutoGameDesign (project no: 630665) and the Horizon 2020 project COM n PLAY SCIENCE (project no: 787476).
	
\bibliographystyle{IEEEtran}
\bibliography{surprise_qd}

\begin{IEEEbiography}[{\includegraphics[width=1in,height=1.25in,clip,keepaspectratio]{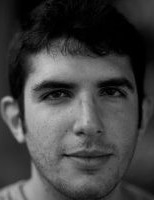}}]
{Daniele Gravina} is a PhD student at the Institute of Digital Games, University of Malta. 
He received the M.Sc. degree in Computer Science from the Polytechnic University of Milan in 2015. His main research interests lie in the areas of computational intelligence, evolutionary computation and automated game design.  
In his doctoral work he focuses on divergent search and quality diversity by means of surprise and introduced a new evolutionary algorithm, named surprise search.

\end{IEEEbiography}

\begin{IEEEbiography}[{\includegraphics[width=1in,height=1.25in,clip,keepaspectratio]{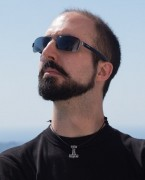}}]{Antonios Liapis} is a Lecturer at the Institute of Digital Games, University of Malta, where he bridges the gap between game technology and game design in courses focusing on human-computer creativity, digital prototyping and game development. He received the Ph.D. degree in Information Technology
from the IT University of Copenhagen in 2014. His research focuses on Artificial Intelligence as an autonomous creator or as a facilitator of human creativity. His work includes computationally intelligent tools for game design, as well as computational creators that blend semantics, visuals, sound, plot and level structure to create horror games, adventure games and more. 
He has received several awards for his research contributions and reviewing effort. He is a member of the IEEE.
\end{IEEEbiography}
\vfill
\begin{IEEEbiography}[{\includegraphics[width=1in,height=1.25in,clip,keepaspectratio]{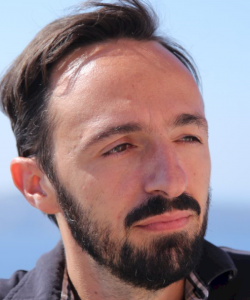}}]{Georgios N. Yannakakis} is a Professor and Director of the Institute of Digital Games, University of Malta. He received the Ph.D. degree in Informatics from the University of Edinburgh in 2006.
He does research at the crossroads of evolutionary computation, computational creativity, affective computing, and game artificial intelligence. He has published more than 220 papers in the aforementioned fields and his work has been cited broadly. His research has been supported by numerous national and European grants (including a Marie Skodowska-Curie Fellowship) and has appeared in \emph{Science Magazine} and \emph{New Scientist}.
He is currently an Associate Editor of the {\sc IEEE Transactions on Games} and used to be Associate Editor of the {\sc IEEE Transactions on Affective Computing} and the {\sc IEEE Transactions on Computational Intelligence and AI in Games} journals.
Among the several rewards he has received for journal and conference publications he is the recipient of the \emph{IEEE Transactions on Affective Computing Most Influential Paper Award} and the \emph{IEEE Transactions on Games Outstanding Paper Award}.  He is a senior member of the IEEE.
\end{IEEEbiography}
\vfill

\end{document}